\documentclass[sigconf]{acmart}
\AtBeginDocument{%
  }

\copyrightyear{2025}
\acmYear{2025}
\setcopyright{licensedothergov}
\acmConference[KDD '25]{Proceedings of the 31st ACM SIGKDD Conference on Knowledge Discovery and Data Mining V.1}{August 3--7, 2025}{Toronto, ON, Canada}
\acmBooktitle{Proceedings of the 31st ACM SIGKDD Conference on Knowledge Discovery and Data Mining V.1 (KDD '25), August 3--7, 2025, Toronto, ON, Canada}
\acmDOI{10.1145/3690624.3709168}
\acmISBN{979-8-4007-1245-6/25/08}

\usepackage{bm}
\usepackage{url}
\usepackage{svg}
\usepackage{tikz}
\usepackage{xcolor}
\usepackage{wrapfig}
\usepackage{amsmath}
\usepackage{caption}
\usepackage{hyperref}
\usepackage{pgfplots}
\usepackage{multirow}
\usepackage{booktabs}
\usepackage{multicol}
\usepackage{graphicx}
\usepackage{placeins}
\usepackage{subfigure}
\usepackage{enumerate}
\usepackage{adjustbox}
\usepackage{algorithm}
\usepackage{algorithmic}
\usepackage{threeparttable}

\begin{document}

%\title{Enhancing Object-Centric Video Learning with Slot-based Spatiotemporal Predictor}
\title{Reasoning-Enhanced Object-Centric Learning for Videos}

\author{Jian Li}
\email{lijian2022@ruc.edu.cn}
\orcid{0000-0002-0685-0861}
\affiliation{%
  \institution{Renmin University of China}
  \city{Beijing}
  \country{China}
}

\author{Pu Ren}
\orcid{0000-0002-6354-385X}
\email{ren.pu@northeastern.edu}
\affiliation{%
  \institution{Northeastern University}
  \city{Boston}
  \state{MA}
  \country{USA}
}

\author{Yang Liu}
\orcid{0000-0003-0127-4030}
\email{liuyang22@ucas.ac.cn}
\affiliation{%
  \institution{University of Chinese Academy of Sciences}
  \city{Beijing}
  \country{China}
}

\author{Hao Sun}
\orcid{0000-0002-5145-3259}
\authornote{Corresponding author.}
\email{haosun@ruc.edu.cn}
\affiliation{%
  \institution{Renmin University of China}
  \city{Beijing}
  \country{China}
}

\renewcommand{\shortauthors}{Jian Li, Pu Ren, Yang Liu and Hao Sun}

\begin{abstract}
Object-centric learning aims to break down complex visual scenes into more manageable object representations, enhancing the understanding and reasoning abilities of machine learning systems toward the physical world. Recently, slot-based video models have demonstrated remarkable proficiency in segmenting and tracking objects, but they overlook the importance of the effective reasoning module. In the real world, reasoning and predictive abilities play a crucial role in human perception and object tracking; in particular, these abilities are closely related to human intuitive physics. Inspired by this, we designed a novel reasoning module called the Slot-based Time-Space Transformer with Memory buffer (STATM) to enhance the model's perception ability in complex scenes. The memory buffer primarily serves as storage for slot information from upstream modules, the Slot-based Time-Space Transformer makes predictions through slot-based spatiotemporal attention computations and fusion. Our experimental results on various datasets indicate that the STATM module can significantly enhance the capabilities of multiple state-of-the-art object-centric learning models for video. Moreover, as a predictive model, the STATM module also performs well in downstream prediction and Visual Question Answering (VQA) tasks. We will release our codes and data at \textit{\url{https://github.com/intell-sci-comput/STATM}}.
\end{abstract}





\maketitle

\section{Introduction}
Objects are the fundamental elements that constitute our world, which adhere to the fundamental laws of physics. Humans learn through activities such as observing the world and interacting with it. They utilize the knowledge acquired via these processes for reasoning and prediction. All these aspects are crucial components of human intuitive physics \cite{lake2017building, kubricht2017intuitive, riochet2018intphys, smith2019promise}. Therefore, object-centric research is pivotal for comprehending human cognitive processes and for developing more intelligent artificial intelligence (AI) systems. By studying the properties, movements, interactions, and behaviors of objects, we can uncover the ways and patterns in which humans think and make decisions in the domains of perception, learning, decision-making, and planning. This contributes to the advancement of more sophisticated machine learning algorithms and AI systems, enabling them to better understand and emulate human intuitive physical abilities \cite{janner2019reasoning, tang2023intrinsic}. 

\begin{figure*}[t!]
\begin{center}
\centerline{\includegraphics[width=0.94\linewidth]{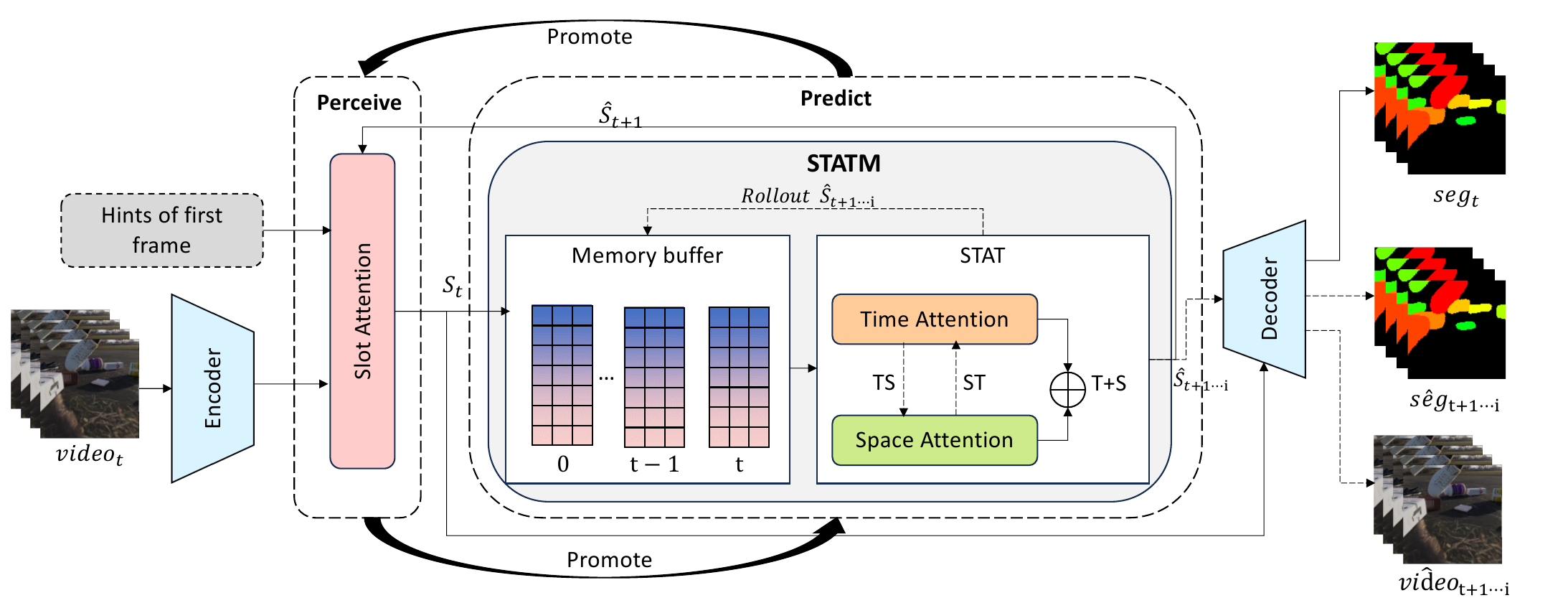}}
\vspace{-6pt}
\caption{Slot-based Time-Space Transformer with Memory buffer architecture overview. The model employs Slot Attention \cite{locatello2020object} for perception, which utilizes slot information predicted by STATM predictor from previous timestep and features extracted by encoder to update slot information. For the first frame, the initial slot information is obtained through either Gaussian distribution or hints module. The updated slot information is then stored in a memory buffer for subsequent use by the TATM. TATM performs reasoning by incorporating temporal cross-attention and spatial self-attention. The integration of temporal and spatial attention can be achieved in various ways. STATM supports both single-step predictions and long-sequence rollouts, where single-step prediction results can be used by Slot Attention to update slot information, and long-sequence rollout results can be used to downstream tasks such as VQA. Both perceptual and predicted slot information can be used by the decoder to obtain reconstruction results and segmentation masks. The architecture features perception and prediction modules that mutually enhance each other.} 
\label{fig1}
\vspace{-8pt}
\end{center}
\end{figure*}

Recently, the representative SAVi \cite{kipf2021conditional} and SAVi++ \cite{elsayed2022SAVi++} models have demonstrated impressive performance in object perception. SAVi (Slot Attention for Video) employed optical flow as a prediction target and leveraged a small set of abstract hints as conditional inputs in the first frame to acquire object-centric representations of dynamic scenes. SAVi++ (Towards End-to-End Object-Centric Learning from Real-World Videos) enhanced the SAVi by integrating depth prediction and implementing optimal strategies for architectural design and data augmentation. Both SAVi and SAVi++ execute two steps on observed video frames: a prediction step and a correction step. The correction step uses inputs to update the slots. The prediction step uses the slots information of the objects provided by the correction step for prediction. The predictor's output initializes the correction process in the subsequent time step, ensuring the model's consistent ability to track objects over time.

The two main steps of such a model operate in a positive feedback loop. The more accurate the predictions, the better the corrections become. Consequently, the more accurate the corrections, the more precise the information provided for the prediction step is, leading to better predictions. Therefore, having a reasonable and efficient predictor is crucial for the entire model. In real-world scenarios, humans also engage in prediction as a crucial aspect of their object perception and tracking, but their prediction behaviors often involve more intricate processes. Humans typically combine the motion state of an object with the interactions of other objects to predict possible future states and positions of the object. The object's motion state is inferred by humans using their common sense from the object's past positions over a while. In so doing, humans enhance their ability to recognize and track relevant objects within complex scenes, which is an integral component of human intuitive physics \cite{sudderth2006graphical, ullman2017mind, mitko2020all}. The prediction step in SAVi and SAVi++ is similar to human inference, but their predictor module is somewhat simplistic, as it relies solely on single-frame information from the current time step for prediction.

In the field of object-centric video prediction, SlotFormer \cite{wu2023SlotFormer} and OCVP  \cite{villar2023object} transform the spatiotemporal attention structure used for video classification in TimeSformer \cite{bertasius2021space} into a similar structure utilized for slot prediction. However, the number of slots (also considered as tokens) required by SlotFormer is likely to increase dramatically with time as the number of objects in the scene grows, leading to an excess of superfluous slot computation. OCVP explored two types of spatiotemporal slot models, which, to some extent, mitigated the increase in token numbers, yet still faced the issue of unnecessary slot computations. The predictive capability of both approaches heavily depends on the quality of the upstream slot extraction. Neither approach made improvements to the upstream module responsible for slot extraction, nor did they delve into the impact of prediction on upstream perception.

Human perception and prediction are typically complementary. When assessing an object's movement, humans often rely on short-term memory impressions of the object. These impressions, along with consideration of environmental factors within the scene, are used to predict the object's movement. Through a comprehensive analysis of time and space, humans anticipate the object's next position. Drawing inspiration from human behavior, we introduce a novel prediction module aimed at enhancing slot-based models for video. This module comprises two key components: 
1) \textbf{Slot-based Memory Buffer}: designed to store historical slot information obtained from the upstream module.
2) \textbf{Slot-based Time-Space Transformer Module}: designed by applying spatiotemporal attention mechanisms to slots for inferring the temporal motion states of objects and calculating spatial objects interactions, which also integrates time and space attention results. The module only computes using slots from the current moment and those from past moments. This not only addresses the issue of the increasing number of tokens as time progresses and the number of objects in the scene increases, but it also reduces unnecessary slot computations. 
We term the proposed model as \textit{\textbf{S}lot-based \textbf{T}ime-Sp\textbf{a}ce \textbf{T}ransformer with \textbf{M}emory buffer} (STATM). Upon substituting the prediction module of SAVi and SAVi++ into the STATM, we observe distinct impacts of different spatiotemporal fusion methods on SAVi and SAVi++. By employing an appropriate fusion method and memory buffer sizes, we observed a significant enhancement in the object segmentation and tracking capabilities of SAVi and SAVi++ on videos containing complex backgrounds and multiple objects per scene. 

Overall, our contributions are summarized as follows:
\begin{itemize}
\item We have investigated the impact of prediction modules on models utilizing Slot Attention \cite{locatello2020object} for object-centric learning from video and have developed the STATM module as a predictor. By simply incorporating a memory buffer and spatiotemporal attention, we have significantly enhanced the capabilities of models like SAVi \cite{kipf2021conditional}, and SAVi++\cite{elsayed2022SAVi++}.
\item We have diligently worked to reduce the computational cost of the spatiotemporal module. In contrast to other models that utilize multiple frames for prediction \cite{wu2023SlotFormer, villar2023object}, our spatiotemporal module only combines current and past slots for computation, effectively decreasing the number of tokens required for the prediction module and the amount of slot computations it performs. As time progresses and the number of objects in the scene increases, the advantage of STATM becomes even more apparent.
\item We have conducted experiments across multiple benchmarks. We have observed that the STATM module significantly enhances the capabilities of multiple state-of-the-art object-centric learning models for video, such as SAVi \cite{kipf2021conditional} , SAVi++ \cite{elsayed2022SAVi++}, and STEVE \cite{singh2022simple}. Moreover, as a predictive model, the STATM also performs well in downstream prediction and Visual Question Answering (VQA) tasks.
% We observe that the integration of the STATM module significantly enhances the segmentation and tracking capabilities of models such as SAVi and SAVi++, particularly on complex datasets. The module notably improves the ability of SAVi to handle complex scenarios and addresses the overfitting issues encountered by SAVi++ on simple datasets. Compared to the original model, the enhanced models just require less training to achieve commendable performance.
\item Furthermore, we have briefly explored the impact of various spatiotemporal architectures and different memory buffer sizes on model performance.
\end{itemize}

\section{Related Work}

\textbf{Object-centric Learning.} In recent years, object-centric learning has emerged as a significant research direction in computer vision and machine learning. It aims to enable machines to perceive and understand the environment from an object-centered perspective, thereby constructing more intelligent visual systems. There is a rich literature on this research, including SQAIR \cite{kosiorek2018sequential}, R-SQAIR \cite{stanic2019r}, SCALOR \cite{jiang2019scalor}, Monet \cite{burgess2019monet}, OP3 \cite{veerapaneni2020entity}, ViMON \cite{weis2020unmasking}, PSGNet \cite{bear2020learning}, SIMONe \cite{kabra2021simone}, and others \cite{kahneman1992reviewing,kipf2019contrastive, zhang2022object, xie2022segmenting}. Slot-based Models represent a prominent approach within object-centric learning. They achieve this by representing each object in a scene as an individual slot, which is used to store object features and attributes \cite{kumar2020ma, yang2021self, dittadi2021generalization, hassanin2022visual, wu2023SlotFormer}.

\textbf{Slot-based Attention and spatiotemporal Attention.}  Our current work is closely related to slot-based attention and spatiotemporal attention. There are a lot of works related to slot-based attention \cite{locatello2020object, hu2020sas, zoran2021parts, ye2021slot, wang2023slot, wu2023SlotFormer}. Spatiotemporal attention mechanisms are particularly effective in handling video data or time-series data, allowing networks to understand and leverage relationships between different time steps or spatial positions \cite{li2020spatio, bertasius2021space, luo2021stan}. Currently, they find wide applications in various fields such as video object detection and tracking \cite{lin2021vehicle, chen2022detecting}, action recognition \cite{yang2022predicting}, natural language processing \cite{xu2020end, weld2022survey}, medical image processing \cite{zhang2020ecg}, among many others \cite{ding2020interpretable, yuan2020deep, cheng2020spatio}.

\textbf{Prediction and Inference on Physics.}  The implementation of object-centric physical reasoning is crucial for human intelligence and is also a key objective in artificial intelligence. Interaction Network \cite{battaglia2016interaction} as the first general-purpose learnable physics engine, is capable of performing reasoning tasks centered around objects or relationships. Another similar study is the Neural Physics Engine \cite{chang2016compositional}. On the other hand, Visual Interaction Networks \cite{watters2017visual} can learn physical laws from videos to predict the future states of objects. Additionally, there are many models developed based on this foundation \cite{chen2021roots, jusup2022social, meng2022physics, piloto2022intuitive, singh2022simple, driess2023learning}. Additionally, there are many object-centric predictive models that are based on slot representations \cite{ding2021dynamic, ding2021attention, wu2023SlotFormer}. However, their performance largely depends on the quality of slot extraction by upstream perception modules. In order to achieve a deeper understanding of commonsense intuitive physics within artificial intelligence systems,  \cite{piloto2022intuitive} have built a system capable of learning various physical concepts, albeit requiring access to privileged information such as segmentation. Our research primarily aims to construct an object-centric system for object perception, learning of physics, and reasoning.

\section{Slot-based Time-Space Transformer with Memory Buffer (STATM)}

To enhance the slot-based video models, we introduce a new module called \textit{Slot-based Time-Space Transformer with Memory Buffer} (STATM) as the predictor. This module consists of two key components: 1) memory buffer, and 2) Slot-based Time-Space Transformer (STAT). The memory buffer serves as a repository for storing historical slot information obtained from upstream modules, while STAT utilizes the information stored in the memory buffer for prediction and causal reasoning. The overall framework is shown in Figure~\ref{fig1}.

\subsection{Memory Buffer}

The memory module is utilized for storing slot information from the upstream modules. We employ a queue-based storage mechanism. The representation of the memory buffer at time $t$ is given by:
\begin{equation}
\label{eq1}
M_t=Queue(S_i, \dots, S_t) ,
\end{equation}
where $S_t= \{s_{(0,t)}, \dots, s_{(N,t)}\}$ represents the slot information extracted from the corrector module at time $t$. Here, $N$ signifies the number of slots, which is associated with the number of objects within the scene. The size of $M$ can be fixed or infinite. The new information is appended at the end of the queue.

\subsection{Slot-based Time-Space Transformer}

The primary role of STAT (Slot-based Time-Space Transformer) lies in leveraging slot data from the memory buffer to facilitate slot-based dynamic temporal reasoning and spatial interaction computations. Furthermore, it integrates the outcomes of temporal reasoning and spatial interactions to achieve a unified understanding. Specifically, for temporal dynamic reasoning, a cross-attention mechanism is employed, which effectively utilizes historical context stored in the memory buffer to enable accurate predictions of future states. Meanwhile, for spatial interaction computations, we employ a self-attention mechanism that operates on slot representations to compute the relevance between different slots within the $S$. The results obtained from temporal dynamic reasoning and spatial interaction computation are merged to provide a holistic understanding encompassing both temporal dynamics and spatial interactions. This comprehensive representation enhances the model's capability for accurate prediction and reasoning in object-centric tasks. We propose three approaches: 
\begin{subequations}
\label{eq7}
\begin{align}
\widehat{S}_{t+1} &= CrossAtt^{time}(S_t,M_t)+SelfAtt^{space}(S_t)\\
\widehat{S}_{t+1} &= CrossAtt^{time}(SelfAtt^{space}(S_t),M_t)\\
\widehat{S}_{t+1} &= SelfAtt^{space}(CrossAtt^{time}(S_t,M_t)) ,
\end{align}
\end{subequations}

(2a) \textit{T+S}: The sum of computed temporal attention and spatial attention. (2b) \textit{ST}: Spatial attention computation followed by using the outcome as input for temporal attention. (2c) \textit{TS}: Temporal attention computation followed by using the outcome as input for spatial attention.

\begin{figure}[t!]
\begin{center}
\centerline{\includegraphics[width=0.98\linewidth]{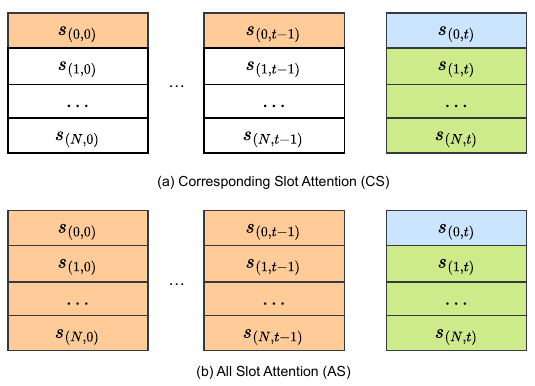}}
\vspace{-6pt}
\caption{Spatiotemporal attention computation architectures. The green slots represent those employed for spatial attention computation, while the orange slots are indicative of those used for temporal attention computation.}
\vspace{-10pt}
\label{fig2}
\vspace{-3pt}
\end{center}
\end{figure}

As shown in Figure \ref{fig2}, we introduce two computational architectures for spatiotemporal attention: (a) \textit{Corresponding Slot Attention (CS)}: For slot $s_{(i,t)}$, temporal attention is computed by using it and corresponding slots in $\{s_{(i,0)}, \dots, s_{(i,t-1)}\}$, while spatial attention computation is performed using it and all slots within $\{s_{(0,t)}, \dots, s_{(N,t)}\}$. (b) \textit{All Slot Attention (AS)}: For slot $s_{(i,t)}$, temporal attention is computed by using it and all slots in $\{s_{(0,0)}, \dots, s_{(N,t-1)}\}$. The spatial attention computation remains the same as in the CS.

In the CS architecture, $s_{(i,t)}$ undergoes temporal attention computation exclusively with its corresponding slots. This design offers several notable advantages. Firstly, it enables a more robust association between objects and slots in terms of temporal sequences, preserving the slot's invariance with respect to the object. Additionally, this approach significantly reduces computational costs when compared to the AS structure. This efficiency makes the CS architecture an appealing choice for achieving effective temporal binding while optimizing computational resources.

In the AS architecture, the temporal attention involves calculating the attention between $s_{(i,t)}$ and all previous slots. 

\subsection{Differences between STATM and Other Slot-based Models}

In the field of object-centric video prediction, SlotFormer \cite{wu2023SlotFormer} and OCVP  \cite{villar2023object} transform the spatiotemporal attention structure used for video classification in TimeSformer \cite{bertasius2021space} into a similar structure utilized for slot prediction. Nevertheless, the computational slots utilized by these models are likely to surge dramatically over time as the number of objects in the scene increases, leading to an excess of unnecessary slot computations.

Our model design is focused on reducing the computational cost. Traditionally, slot-based models, such as SlotFormer\cite{wu2023SlotFormer}, flatten the time $T$ ($T>2$) and the number of slots $N$ ($N>2$) when using $T$ time steps for prediction, then input them into the transformer encoder to calculate full attention between each slot. Thus, the transformer encoder must process $T\times N$ tokens and perform attention calculations on $T\times N$ slots, leading to $(T\times N)^2$ calculations. 

In contrast, our spatiotemporal modules only combine the current time slots with the previous time slots for calculation. Therefore, the prediction modules based on the CS structure only requires $T+N$ tokens, and the AS structure requires $T\times N$ tokens. In attention calculations, we only consider computations between the current moment slots and other slots. For the CS structure, in terms of time attention, it only needs to calculate attention between current time slot $s_i$ ($i=1,\dots,N$) and the other $T-1$ slots, and in terms of space attention, it only needs to calculate attention between $N$ slots, thus involving $N\times (T-1) + N^2$ slot calculations in total. For the AS structure, it involves $(T-1) \times N \times N$ calculations in time and $N^2$ calculations in space, totaling $T\times N^2$ calculations. Clearly, aside from the token requirement for the AS structure, both the number of tokens and the amount of computation required by our spatiotemporal attention module are significantly less than that of SlotFormer. We also found that using only the CS structure can achieve good model performance. As the number of frames and objects in a scene increase, the CS structure is more efficient.

\subsection{Training}

Our focus is on conducting perception and prediction experiments. In the perception experiments, we intend to enhance several state-of-the-art object-centric learning models using the STATM module. Consequently, the training loss for each model may differ. We train STATM-SAVi and STATM-SAVi++ to minimize the L2 loss between the predicted and ground-truth targets, such as optical flow, images, and depth signals. The training loss for STATM-STEVE aligns with the loss function used in STEVE\cite{singh2022simple}.

In the prediction experiments, we train the STATM model by jointly minimizing slots and images reconstruction loss (also L2).

\section{Experiments}

Our experiments primarily comprise three parts: perception, prediction and VQA, and ablation. The perception experiments aim to verify the impact of the STATM module on existing state-of-the-art object-centric learning models for video. The prediction experiments are designed to preliminarily demonstrate the robust performance of the STATM module as a predictive model in downstream prediction and VQA tasks. The ablation studies focus on assessing the effects of the memory buffer and various spatiotemporal structures on model performance.

\textbf{Baselines.} In the perception experiments, we primarily compare SAVi \cite{kipf2021conditional}, SAVi++ \cite{elsayed2022SAVi++}, STEVE \cite{singh2022simple} and SAVi-SlotFormer. For SAVi, we chose official implementation, SAVi-small, as the baseline. The SAVi-samll model includes five components: encoder, decoder, slot initialization, corrector, and predictor. The encoder uses a CNN to extract features from video frames. Slot initialization, using either an MLP or a CNN, prepares slots with initial data like bounding boxes. The corrector, powered by Slot Attention \cite{locatello2020object}, updates slots using encoder features. The predictor, a transformer block, uses self-attention for forecasting and initializes the corrector for consistent tracking. Finally, the decoder outputs RGB predictions and an alpha mask using a Spatial Broadcast Decoder. The SAVi-SlotFormer is a baseline we develope to assess the impact of the prediction module on the perceptual performance of SAVi. SAVi++ \cite{elsayed2022SAVi++} has a structure similar to SAVi with a ResNet34 backbone. Unlike SAVi, SAVi++ introduces depth as self-supervised objectives. It also incorporates data augmentation and utilizes a transformer encoder after ResNet34. 

In the prediction section, we compare G-SWM \cite{lin2020improving}, SAVi-dyn\cite{wu2023SlotFormer}, and SlotFormer \cite{wu2023SlotFormer}. SlotFormer \cite{wu2023SlotFormer} is a transformer-based framework for object-centric visual simulation. It leverages slots extracted by upstream modules like SAVi to train a slot-based transformer encoder model for prediction purposes.

For the VQA experiments, our main comparisons involve DCL \cite{chen2021grounding}, VRDP \cite{ding2021dynamic}, and SlotFormer \cite{wu2023SlotFormer}. SlotFormer utilizes prediction results from rollout simulations to train Aloe \cite{ding2021attention} for VQA tasks. VRDP \cite{ding2021dynamic} is designed to jointly learn visual concepts and infer physics models of objects and their interactions from both videos and language. It primarily consists of three modules: a visual perception module, a concept learner, and a differentiable physics engine. We have implemented VRDP with a visual perception module that is trained based on object properties. 

%For more description about baselines, please refer to Appendix Section ~\ref{append-baseline}. 
For more baselines, please refer to Appendix Section ~\ref{append-baseline}. 

\textbf{Metrics.} To evaluate the model's object-centric learning capability for video, we selected the Adjusted Rand Index (ARI) \cite{rand1971objective, hubert1985comparing} and the mean Intersection over Union (mIoU) as evaluation metrics. ARI quantifies the alignment between predicted and ground-truth segmentation masks. For scene decomposition assessment, we commonly employ FG-ARI and, which is a permutation-invariant clustering similarity metric. It allows us to compare inferred segmentation masks to ground-truth masks while excluding background pixels. mIoU is a widely used segmentation metric that calculates the mean Intersection over Union values for different classes or objects in a segmentation task. To assess video quality, we report PSNR, SSIM \cite{wang2004image}, and LPIPS \cite{zhang2018unreasonable}. To evaluate the prediction outcomes, we utilize FG-ARI and FG-mIoU.

\textbf{Datasets.} To evaluate the object-centric learning capability, we utilized the synthetic Multi-Object Video (MOVi) datasets \cite{googleresearch2020, greff2022kubric}. These datasets are divided into five distinct categories: A, B, C, D, and E. MOVi-A and B depict relatively straightforward scenes, each containing a maximum of 10 objects. MOVi-C, D, and E present more intricate scenarios with complex natural backgrounds. MOVi-C, generated using a stationary camera, presents scenes with up to 10 objects. Transitioning to MOVi-D, the dataset extends the object count to accommodate a maximum of 23 objects. Lastly, MOVi-E introduces an additional layer of complexity by incorporating random linear camera movements. Each video sequence is sampled at a frame rate of 12, resulting in a total of 24 frames per second.

To assess the predictive and Visual Question Answering (VQA) capabilities, we have selected the CLEVRER \cite{yi2019CLEVRER} dataset. CLEVRER dataset is specifically designed for video understanding and reasoning, focusing on the dynamics of objects and their causal interactions. For the VQA task, CLEVRER incorporates four types of questions: descriptive (e.g., "what color"), explanatory ("what's responsible for"), predictive ("what will happen next"), and counterfactual ("what if"). The predictive questions require the model to simulate future object interactions, such as collisions. Thus, we are particularly concentrating on enhancing the accuracy of predictive questions through the implementation of STATM's future rollout.

\textbf{Training Setup.} In all experiments except the ablation study in Section~\ref{exp-43}, we used the STAT encoding block in combination with the CS attention architecture, featuring the T+S spatiotemporal fusion approach. For perception experiments, we utilized videos comprising of 6 frames at a resolution of 64$\times$64 pixels to train the STATM-SAVi and SAVi models. The training process is conducted over 100k iterations. Similarly, the STATM-SAVi++ and SAVi++ models were trained on continuous videos consisting of 6 frames at a higher resolution of 128$\times$128 pixels, with training duration encompassing 100k or 500k iterations. The buffer size was unconstrained during training, and the maximum length of effective information was limited to 6 due to the utilization of a 6-frame training sequence. Bounding boxes were used as the conditioning for all models. For prediction and VQA experiments, we train our models (STATM-SAVi) for 400k steps with a batch size of 64 on the CLEVRER dataset to extract slots. The number of slots is set to 7, with a learning rate of 0.0001. We subsample the video by a factor of 2 to train STATM, conducting approximately 500k training steps with a batch size of 64 and a learning rate of 0.0002. We use rollout slots to train Aloe \cite{ding2021attention}, targeting around 300k steps with a learning rate of 0.0001 and a batch size of 128. We use the Adam optimizer and apply warm-up and decay learning rate schedule for the first 2.5\% of the total training steps. For more training setup, please refer to Appendix Section ~\ref{append-baseline} and ~\ref{append-trainsetup}. 

\subsection{Perception} 
\label{exp-41}

\begingroup
\setlength{\tabcolsep}{0.18em}
\begin{table*}[t!]
\renewcommand\arraystretch{1.1}
\caption{Enhancement results by STATM on models with hints. The first five rows depict the evaluation results for models trained for 100k steps with a batch size of 32. SAVi-SlotFormer$^{*}$ denotes our implemented  baseline model. SAVi++$^{*}$ represents results from SAVi++ paper \cite{elsayed2022SAVi++}. STATM-SAVi++$^{*}$ denotes the evaluation results for STATM-SAVi++ model trained for 500k steps with a batch size of 64 (Mean $\pm$ standard error over 3 seeds). }
\vspace{-6pt}
\label{tab1}
\begin{center}
\begin{tabular}{lccccclccccc}
\toprule
\multicolumn{1}{l}{\bf \multirow{2}*{Model}} &\multicolumn{5}{c}{\bf mIoU$\uparrow(\%)$} &\multicolumn{1}{l}{} &\multicolumn{5}{c}{\bf FG-ARI$\uparrow(\%)$} \\
\cline{2-6} \cline{8-12} 
{} &A&B&C&D&E&{}&A&B&C&D&E \\
\hline 
SAVi & 62.8&41.6&22.0&6.8&4.0&{}&\bm{$91.1$}&70.2&50.4&18.4&10.8 \\
SAVi-SlotFormer$^{*}$ &63.5&-&-&-&7.5&{}&86.4&-&-&-&31.2 \\
STATM-SAVi (Ours) &\bm{$67.5$}&\bm{$42.8$}&\bm{$34.0$}&\bm{$17.0$}&\bm{$9.0$}&{}&\bm{$91.1$}&\bm{$70.7$}&\bm{$57.7$}&\bm{$40.9$}&\bm{$36.9$} \\
\cline{1-12}
SAVi++ & 82.8&\bm{$52.5$}&47.8&43.6&26.1&{}&96.7&78.5&76.3&81.5&81.7 \\
STATM-SAVi++ (Ours) &\bm{$83.5$}&\bm{$52.5$}&\bm{$49.5$}&\bm{$50.1$}&\bm{$27.9$}&{}&\bm{$96.9$}&\bm{$78.9$}&\bm{$77.7$}&\bm{$85.8$}&\bm{$85.0$} \\
\cline{1-12}
SAVi++$^{*}$ & {$76.1{\pm0.9}$}&{$25.8{\pm11.3}$}&{$45.2{\pm0.1}$}&{$48.3{\pm0.5}$}&{$47.1{\pm1.3}$}&{}&{$98.2{\pm0.2}$}&{$48.3{\pm15.7}$}&{$81.9{\pm0.2}$}&{$86.0{\pm0.3}$}&{$84.1{\pm0.9}$} \\
STATM-SAVi++$^{*}$ (Ours) & {$\bm{85.6}{\pm0.6}$}&{$\bm{60.4}{\pm1.2}$}&{$\bm{52.4}{\pm0.2}$}&{$\bm{57.0}{\pm0.4}$}&{$\bm{55.4}{\pm0.9}$}&{}&{$\bm{98.3}{\pm0.2}$}&{$\bm{84.9}{\pm2.5}$}&{$\bm{82.2}{\pm0.2}$}&{$\bm{89.1}{\pm0.2}$}&{$\bm{88.6}{\pm0.5}$} \\
\bottomrule
\end{tabular}
\end{center}
\end{table*}
\endgroup

\begin{figure*}[t!]
\vspace{3pt}
\begin{small}
    \centering    
    \begin{minipage}[t]{1.0\linewidth}
    \centering
        \begin{tabular}{@{\extracolsep{\fill}}c@{}c@{}c@{}@{\extracolsep{\fill}}}
        MOVi-A \hspace{4.2cm} MOVi-B \hspace{4.2cm} MOVi-D
        \end{tabular}
    \end{minipage}
    \begin{minipage}[t]{1.0\linewidth}
    \centering
    \rotatebox[origin=c]{90}{Video} \
        \begin{tabular}{@{\extracolsep{\fill}}c@{}c@{}c@{}@{\extracolsep{\fill}}}
            \includegraphics[width=0.32\linewidth]{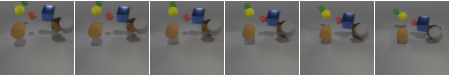}& \        \includegraphics[width=0.32\linewidth]{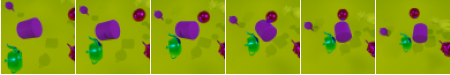}& \   
            \includegraphics[width=0.32\linewidth]{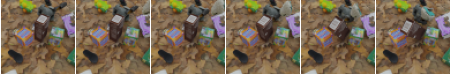}\\
        \end{tabular}
    \end{minipage}
    \begin{minipage}[t]{1.0\linewidth}
    \centering
    \rotatebox[origin=c]{90}{G.T.} \
        \begin{tabular}{@{\extracolsep{\fill}}c@{}c@{}c@{}@{\extracolsep{\fill}}}
            \includegraphics[width=0.32\linewidth]{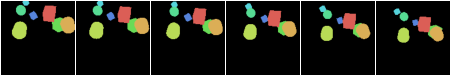}& \        \includegraphics[width=0.32\linewidth]{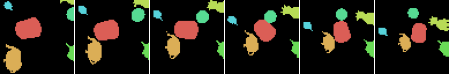}& \  
            \includegraphics[width=0.32\linewidth]{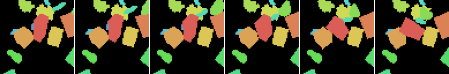}\\
        \end{tabular}
    \end{minipage}   
    \begin{minipage}[t]{1.0\linewidth}
    \centering
    \rotatebox[origin=c]{90}{SAVi} \
        \begin{tabular}{@{\extracolsep{\fill}}c@{}c@{}c@{}@{\extracolsep{\fill}}}
            \includegraphics[width=0.32\linewidth]{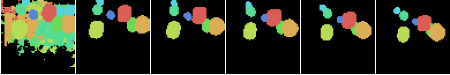}& \        \includegraphics[width=0.32\linewidth]{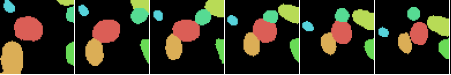}& \   
            \includegraphics[width=0.32\linewidth]{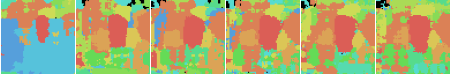}\\
        \end{tabular}
    \end{minipage}
    \begin{minipage}[t]{1.0\linewidth}
    \centering
    \rotatebox[origin=c]{90}{\textbf{Ours}} \
        \begin{tabular}{@{\extracolsep{\fill}}c@{}c@{}c@{}@{\extracolsep{\fill}}}
            \includegraphics[width=0.32\linewidth]{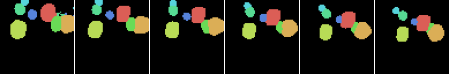}& \        \includegraphics[width=0.32\linewidth]{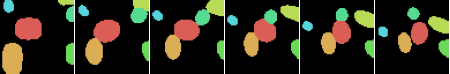}& \  
            \includegraphics[width=0.32\linewidth]{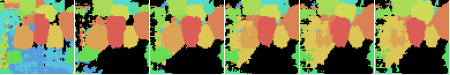}\\
        \end{tabular}
    \end{minipage}   
    \begin{minipage}[t]{1.0\linewidth}
    \rotatebox[origin=c]{90}{SAVi++} \
    \centering
        \begin{tabular}{@{\extracolsep{\fill}}c@{}c@{}c@{}@{\extracolsep{\fill}}}
            \includegraphics[width=0.32\linewidth]{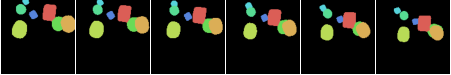}& \        \includegraphics[width=0.32\linewidth]{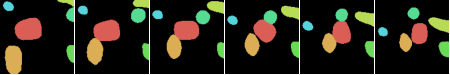}& \   
            \includegraphics[width=0.32\linewidth]{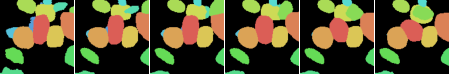}\\
        \end{tabular}
    \end{minipage}
    \begin{minipage}[t]{1.0\linewidth}
    \centering
    \rotatebox[origin=c]{90}{\textbf{Ours++}} \
        \begin{tabular}{@{\extracolsep{\fill}}c@{}c@{}c@{}@{\extracolsep{\fill}}}
            \includegraphics[width=0.32\linewidth]{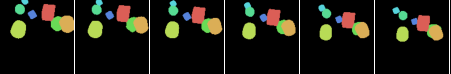}& \        \includegraphics[width=0.32\linewidth]{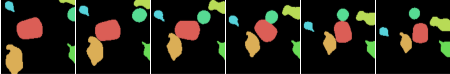}& \  
            \includegraphics[width=0.32\linewidth]{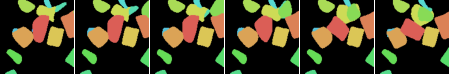}\\
        \end{tabular}
    \end{minipage}   
    \vspace{-6pt}
    \caption{Qualitative results of our model compared to SAVi and SAVi++ on the MOVi dataset. Compared with SAVi and SAVi++, our model is slightly better than the SAVi/SAVi++ mode on the relatively simple datasets. As the complexity of the datasets increases, the advantage of our model becomes more pronounced.}
\label{fig3}
%\vspace{-6pt}
\end{small}
\end{figure*}

\begin{table}[t!]
\renewcommand\arraystretch{1.1}
\caption{Enhancement Results (FG-ARI\%) on STEVE.}
\vspace{-6pt}
\label{tab-s1}
\begin{center}
\begin{tabular}{lcc}
\toprule
\multicolumn{1}{c}{\bf Model} &\multicolumn{1}{c}{\bf MOVi-D} &\multicolumn{1}{c}{\bf MOVi-E} \\
\hline 
STEVE & 47.67 & 52.15 \\
STATM-STEVE (Ours) & $\bm{51.73}$ & $\bm{55.78}$ \\
\bottomrule
\end{tabular}
\vspace{-6pt}
\end{center}
\end{table}

\textbf{Results.} Quantitative results can be seen in Table~\ref{tab1} and ~\ref{tab-s1}, and qualitative results in Figure~\ref{fig3}. From Table~\ref{tab1} and ~\ref{tab-s1}, it is evident that STATM can significantly enhance the performance of existing state-of-the-art models. Comparing SAVi, SAVi-SlotFormer and STATM-SAVi, SAVi performs reasonably well on simple datasets but struggles with complex datasets. When SlotFormer is used as the predictor, it enhances SAVi's performance on complex datasets, yet it reduces its effectiveness on simpler datasets. Conversely, when STATM serves as the predictor, it not only improves SAVi's performance on complex datasets but also maintains its performance on simple datasets. In comparison between SAVi++ and STATM-SAVi++, the enhanced model shows a notable improvement, and except for the MOVi-E (due to insufficient training), it can generally match or surpass the optimal results of SAVi++. Further, when considering the best results of STATM-SAVi++$^{*}$ against the official results SAVi++, our model's performance is markedly superior to the original model, indicating that the model's benefits do not diminish with increased training. The limitations of depth information and overfitting mentioned in the SAVi++ \cite{elsayed2022SAVi++} do not appear in STATM-SAVi++. The qualitative results from Figure~\ref{fig3} also demonstrate the superior performance of our models (e.g., the notable difference between SAVi and STATM-SAVi in the first frame of MOVi-A). 

In addition, we also tried to use the STATM module to improve the unsupervised scene segmentation model STEVE. The results can be found in Table~\ref{tab-s1}. We found that STAM remains effective for unsupervised object-centric learning model. % in Appendix Section~\ref{append-result}.

\begin{figure*}[!t]
\begin{small}
    \centering    
    \begin{minipage}[t]{1.0\linewidth}
    \centering
        \begin{tabular}{@{\extracolsep{\fill}}c@{}c@{}@{\extracolsep{\fill}}}
        (a) New object appear \qquad\qquad \hspace{4.5cm} (b) Object reappears after being occluded
        \end{tabular}
    \end{minipage}
    \begin{minipage}[t]{1.0\linewidth}
    \centering
    \rotatebox[origin=c]{90}{Video} \
        \begin{tabular}{@{\extracolsep{\fill}}c@{}c@{}@{\extracolsep{\fill}}}
            \includegraphics[width=0.35\linewidth]{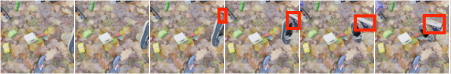}& \ 
            \includegraphics[width=0.6\linewidth]{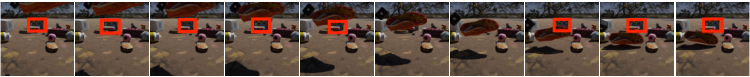}\\
        \end{tabular}
    \end{minipage}
    \begin{minipage}[t]{1.0\linewidth}
    \centering
    \rotatebox[origin=c]{90}{G.T.} \
        \begin{tabular}{@{\extracolsep{\fill}}c@{}c@{}@{\extracolsep{\fill}}}
            \includegraphics[width=0.35\linewidth]{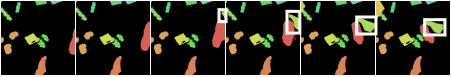}& \
            \includegraphics[width=0.6\linewidth]{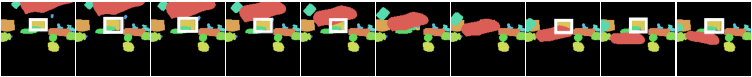}\\
        \end{tabular}
    \end{minipage}   
    \begin{minipage}[t]{1.0\linewidth}
    \rotatebox[origin=c]{90}{SAVi++} \
    \centering
        \begin{tabular}{@{\extracolsep{\fill}}c@{}c@{}@{\extracolsep{\fill}}}
            \includegraphics[width=0.35\linewidth]{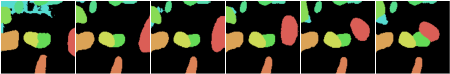}& \ 
            \includegraphics[width=0.6\linewidth]{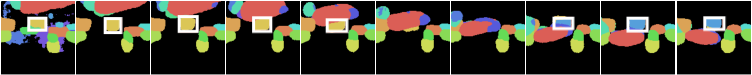}\\
        \end{tabular}
    \end{minipage}
    \begin{minipage}[t]{1.0\linewidth}
    \centering
    \rotatebox[origin=c]{90}{\textbf{Ours++}} \
        \begin{tabular}{@{\extracolsep{\fill}}c@{}c@{}@{\extracolsep{\fill}}}
            \includegraphics[width=0.35\linewidth]{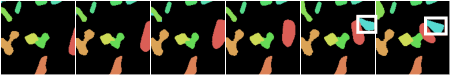}& \ 
            \includegraphics[width=0.6\linewidth]{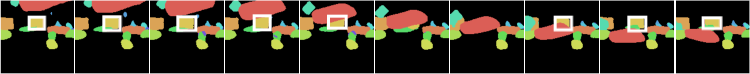}\\
        \end{tabular}
    \end{minipage}   
    % \vspace{3pt} \\
    % \begin{minipage}[t]{1.0\linewidth}
    % \centering
    %     \begin{tabular}{@{\extracolsep{\fill}}c@{}@{\extracolsep{\fill}}}
    %     (c) Object reappears after being occluded
    %     \end{tabular}
    % \end{minipage}
    % \begin{minipage}[t]{1.0\linewidth}
    % \centering
    % \rotatebox[origin=c]{90}{Video} \
    %     \begin{tabular}{@{\extracolsep{\fill}}c@{}@{\extracolsep{\fill}}}
    %     \includegraphics[width=0.95\linewidth]{img/seg1/cover2_gtvideos_11_batch_0.png}\\
    %     \end{tabular}
    % \end{minipage}
    % \begin{minipage}[t]{1.0\linewidth}
    % \centering
    % \rotatebox[origin=c]{90}{G.T.} \
    %     \begin{tabular}{@{\extracolsep{\fill}}c@{}@{\extracolsep{\fill}}}
    %         \includegraphics[width=0.95\linewidth]{img/seg1/cover2_gtsegs_11_batch_0.png}\\
    %     \end{tabular}
    % \end{minipage}   
    % \begin{minipage}[t]{1.0\linewidth}
    % \rotatebox[origin=c]{90}{SAVi++} \
    % \centering
    %     \begin{tabular}{@{\extracolsep{\fill}}c@{}@{\extracolsep{\fill}}}
    %         \includegraphics[width=0.95\linewidth]{img/seg1/cover2_prsegs_11_batch_0_yuan.png}\\
    %     \end{tabular}
    % \end{minipage}
    % \begin{minipage}[t]{1.0\linewidth}
    % \centering
    % \rotatebox[origin=c]{90}{Ours++} \
    %     \begin{tabular}{@{\extracolsep{\fill}}c@{}@{\extracolsep{\fill}}}
    %         \includegraphics[width=0.95\linewidth]{img/seg1/cover2_prsegs_11_batch_0_our.png}\\
    %     \end{tabular}
    % \end{minipage}  
    \vspace{-6pt}
    \caption{Qualitative results of our model compared to SAVi++. (a) When a new object appears, the SAVi++ cannot recognize it, but our model can correctly identifies it after 1-2 frames. (b) When an object reappears after being obscured, the SAVi++ either assigns it to a different slot (color change) or fails to recognize it. In contrast, our model can correctly identify it.}
\label{fig4}
%\vspace{-6pt}
\end{small}
\end{figure*}

From Figure~\ref{fig4}, it is evident that both SAVi and SAVi++ face challenges in recognizing newly appearing objects and objects that reappear after being occluded. When a new object emerges, the original algorithm often misidentifies it as background or an already occupied slot is taken over by the new object. With the incorporation of STATM, although the model may not immediately segment the new object, as the historical information accumulates in the memory buffer, the STAT module can gradually provide the corrector with hints required for the object segmentation (e.g., position or shape), eventually, the model can successfully segment the object. When an object disappears (possibly temporarily occluded), SAVi++ immediately releases the slot associated with the object, potentially causing difficulty in binding the object to the original slot or even failing to recognize it upon reappearance. For STATM-SAVi++, due to the presence of the memory buffer and temporal attention in STATM, when the occluded object reappears, it can easily be assigned to its historical slot. Only when the object has been absent for an extended period will the slot be released.

\textbf{Memory Cost and Inference Time}. The memory cost and inference time (taken to process 250 videos) are listed in Table~\ref{tab3}. We can see that adding STATM to the model doesn't bring much extra memory cost, inference time and parameters. Detailed comparison of parameters can be found in Appendix Table \ref{tab-a1}  Appendix Section~\ref{append-parameter}. 

\begin{table}[t!]
\caption{Memory cost and inference time on MOVi-A and E validation sets, with each set containing 250 videos, each video contains 24 frames (batch size of 32, on one A100 GPU). } 
\label{tab3}
\vspace{-6pt}
\begin{tabular}{lcclcclcc}
\toprule
\multicolumn{1}{c}{\bf \multirow{2}*{Model}} &\multicolumn{2}{c}{\bf Memory (GB)} &\multicolumn{1}{l}{} &\multicolumn{2}{c}{\bf Infer. Time (s)}  \\
\cmidrule{2-3} \cmidrule{5-6} 
{} & A & E & {} & A & E \\
\midrule 
SAVi& 25.91&56.42&{}&78.4&141.3\\
SAVi\small{-SlotFormer}& 25.92&56.43&{}&102.6&204.1\\
\small{STATM-}SAVi (Ours)& 25.91&56.42&{}&81.4&146.4\\
\bottomrule
\end{tabular}
%\vspace{-6pt}
\end{table}

\textbf{Generalization.}We selected the models trained with a batch size of 32 and 100k training steps to assess its generalization. The test sets utilized the default test split of MOVi-C, D and E dataset, featuring scenes exclusively consist of held-out objects and background images to evaluate generalization. The results are presented in Table~\ref{tab-d1} and Figure~\ref{fig5} (a).

\begin{table}[t!]
\renewcommand\arraystretch{1.1}
\caption{Generalization results on MOVi datasets.}
\label{tab-d1}
\vspace{-10pt}
\begin{center}
\begin{tabular}{lccclccc}
\toprule
\multicolumn{1}{l}{\bf \multirow{2}*{Model}} &\multicolumn{3}{c}{\bf mIoU$\uparrow(\%)$} &\multicolumn{1}{l}{} &\multicolumn{3}{c}{\bf FG-ARI$\uparrow(\%)$} \\
\cline{2-4} \cline{6-8} 
{} &C&D&E&{}&C&D&E \\
\hline 
\small SAVi-S (IID) &22.0&6.8&4.0&{}&50.4&18.4&10.8 \\
\small SAVi-S (OOD) &21.1&6.3&3.7&{}&52.8&19.8&9.9 \\
\cline{1-8}
\small STATM-SAVi-S (IID) &34.0&17.0&9.0&{}&57.7&40.9&36.9 \\
\small STATM-SAVi-S (OOD) &33.2&17.0&8.2&{}&59.7&43.4&35.9 \\
\cline{1-8}
\small SAVi++ (IID) &47.8&43.6&26.1&{}&76.3&81.5&81.7 \\
\small SAVi++ (OOD) &46.9&44.0&25.5&{}&77.7&82.2&82.5 \\
\cline{1-8}
\small STATM-SAVi++ (IID) &49.5&50.1&27.9&{}&77.7&85.8&85.0 \\
\small STATM-SAVi++ (OOD) &48.7&50.3&27.4&{}&78.9&86.4&85.8 \\
\bottomrule
\end{tabular}
\vspace{-6pt}
\end{center}
\end{table}

\textbf{Discussion.} Certainly, the STATM module, acting as a predictor, significantly enhances the model performance of SAVi and SAVi++. Compared to the original models, the improved model can achieve good performance with fewer training steps and a smaller batch size. STATM-SAVi++ also addresses the overfitting issue of SAVi++ on simple datasets (especially noticeable in MOVi-B). The enhanced models also exhibit good generalization. Importantly, the integration of a STAT encoding block does not lead to a significant increase in memory cost and inference time. 

\subsection{Prediction and VQA}

In this section, our primary objective is to succinctly validate the performance of STATM in downstream prediction and VQA tasks.

\textbf{Perception on CLEVRER}. Due to the effectiveness of object-centric predictive models largely depends on the quality of slots extracted by upstream perceptual modules. Additionally, MOVi validation sets only includes sequences of 24 frames in length. Therefore, in Table \ref{tab4-1}, we demonstrate a comparison of perceptual effects on CLEVER with longer sequences of 128 frames. It is evident that STATM-SAVi achieves a significantly higher FG-ARI compared to SAVi, indicating that STATM provides a more pronounced enhancement on SAVi in longer sequences on relatively simple datasets.

\begin{table}[t!]
\caption{Perception results on CLEVRER.}
\label{tab4-1}
\vspace{-6pt}
\begin{tabular}{lccc}
\toprule
{\bf Model}&{\bf MSE$\downarrow$}&{\bf FG-ARI}(\%)&{\bf FG-mIoU}(\%)\\
\midrule
SAVi&0.24&91.4&\textbf{77.6}\\
STATM-SAVi (Ours)&\textbf{0.15}&\textbf{93.5}&\textbf{77.6}\\
\bottomrule
\end{tabular}
\vspace{-6pt}
\end{table}

\textbf{Prediction}. Table \ref{tab4-2} displays the evaluation results of visual quality and object dynamics on CLEVRER. It is evident that when using slots extracted by SAVi to train our STATM for prediction, the model shows improvement in all metrics except PSNR, indicating a certain advantage of our model in prediction tasks. When using slots extracted by STATM-SAVi to train STATM for predictions, both video quality and predictive metrics see further enhancements, particularly FG-ARI. This further demonstrates the advantages of our model in both perception and prediction.

\begin{table}[t!]
\caption{Evaluations of visual quality (columns 2-4) and object dynamics (columns 5-6) on CLEVRER. SA+ST and ST+ST respectively represent the results of using SAVi and STATM-SAVi for slot extraction followed by training STATM.}
\label{tab4-2}
%\footnotesize
\vspace{-6pt}
\begin{tabular}{lccccc}
\toprule
{\bf Model}&\small {\bf PSNR}&\small {\bf SSIM}&\small {\bf LPIPS}$\downarrow$&\small {\bf FG-ARI}\textsf{(\%)}&\small {\bf FG-mIoU}\textsf{(\%)}\\
\midrule
SAVi-Dyn&29.77&\textbf{0.89}&0.19&64.32&18.25\\
SlotFormer&30.21&\textbf{0.89}&\textbf{0.11}&63.00&49.40\\
\small SA+ST(Ours)&30.10&\textbf{0.89}&\textbf{0.11}&63.11&49.55\\
\small ST+ST(Ours)&\textbf{30.22}&\textbf{0.89}&\textbf{0.11}&\textbf{64.56}&\textbf{49.57}\\
\bottomrule
\end{tabular}
\vspace{-6pt}
\end{table}

\begin{figure*}[t!]
\begin{center}
\centerline{\includegraphics[width=0.96\linewidth]{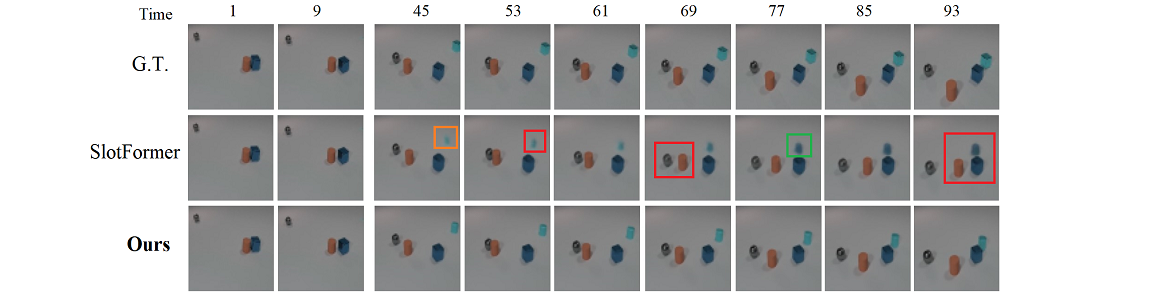}}
\vspace{-6pt}
\caption{Results of long-sequence prediction on CLEVRER. After surpassing a certain time point, results generated by SlotFormer in prediction clearly begin to deviate from the ground truth, exhibiting artifacts such as blurry (orange boxes), incorrect dynamics (red boxes), and inaccurate colors (green boxes). Meanwhile, our model demonstrates good performance. } 
\label{fig-cl}
\vspace{-6pt}
\end{center}
\end{figure*}

Figure \ref{fig-cl} displays generation results of long-sequence prediction on CLEVRER. We observe that SlotFormer performs well in shorter sequence predictions. However, as time progresses, SlotFormer's performance deteriorates significantly, exhibiting blurriness, incorrect dynamics, and inaccurate colors, even becoming completely inconsistent with the ground truth. In contrast, our model continues to perform well.

\textbf{VQA}. In Table \ref{tab4-3}, we present the accuracy on predictive questions. Notably, also as an unsupervised predictive model, our method surpasses the previous state-of-the-art SlotFormer \cite{wu2023SlotFormer}. Furthermore, on the publicly available CLEVRER leaderboard for the predictive question subset, our approach achieved first rank in the per option setting and second rank in the per question setting. %For more detailed results, please refer to Table~\ref{tab-d2} in Appendix Section \ref{append-result}.

\begin{table}[t!]
\caption{Predictive VQA accuracy on CLEVRER.}
\label{tab4-3}
\vspace{-10pt}
\begin{tabular}{lcc}
\toprule
{\bf Model}&{\bf per opt.}(\%)&{\bf per ques.}(\%)\\
\midrule
DCL&90.52&82.03\\
VRDP&95.68&91.35\\
SlotFormer&96.50&93.29 \\
STATM (Ours)&\textbf{96.62}&\textbf{93.63} \\
\bottomrule
\end{tabular}
\vspace{-6pt}
\end{table}

% \begin{table*}[t!]
% \centering
% \small
% \begin{minipage}[c]{0.62\textwidth}
% %\centering
% \caption{}\label{tab4-2}
% \begin{tabular}{lcccccc}
% \toprule
% {\bf Model}&{\bf PSNR}&{\bf SSIM}&{\bf LPIPS}&{\bf FG-ARI} (\%)&{\bf FG-mIoU} (\%)\\
% \midrule
% G-SWM&28.42&\textbf{0.89}&0.16&49.61&24.44\\
% SAVi-Dyn&29.77&\textbf{0.89}&0.19&64.32&18.25\\
% SlotFormer&30.21&\textbf{0.89}&\textbf{0.11}&63.00&49.40\\
% SAVi+ST(Ours)&30.10&\textbf{0.89}&\textbf{0.11}&63.11&49.55\\
% ST+ST(Ours)&\textbf{30.22}&\textbf{0.89}&\textbf{0.11}&\textbf{64.56}&\textbf{49.57}\\
% \bottomrule
% \end{tabular}
% \end{minipage}
% \begin{minipage}[c]{0.37\textwidth}
% %\centering
% \caption{VQA results on the CLEVRER dataset.}\label{tab4-3}
% \begin{tabular}{lcc}
% \toprule
% {\bf Model}&{\bf per opt.}(\%)&{\bf per ques.}(\%)\\
% \midrule
% DCL&90.52&82.03\\
% VRDP&95.68&91.35\\
% SlotFormer&96.50&93.29 \\
% STATM (Ours)&\textbf{96.62}&\textbf{93.63} \\
% \bottomrule
% \end{tabular}
% \end{minipage}
% \end{table*}

\textbf{Discussion.} Clearly, our model demonstrates distinct advantages in perception over longer sequences, as well as in downstream prediction and VQA tasks.

\subsection{Ablation Study}
\label{exp-43}

\begin{figure*}[t!]
\centering
\begin{tikzpicture}
\begin{axis}[
ybar,
title={(a) OOD generalization},
xlabel={Dataset},
ylabel={FG-ARI$(\%)$},
xtick={1,2,3}, 
xticklabels={MOVi-C, MOVi-D, MOVi-E}, 
width=0.31\textwidth,
height=4.2cm,
y label style={at={(axis description cs:0.2,0.5)},anchor=south},
ymin=0, ymax=100,
axis lines=left,
enlarge x limits={abs=0.5},
legend pos=north west,
legend style={
    draw=none,
    legend image code/.code={\fill[draw=black,fill=white] (0cm,-0.1cm) rectangle (0.4cm,0.2cm);},
    legend columns=2,  
    at={(0.2,1.08)},
},
] 
%\addplot+ coordinates {(1,49.5) (2,50.1) (3,27.9)};  Miou-IID
%\addplot+ coordinates {(1,48.7) (2,50.3) (3,27.4)};  Miou-OOD
\addplot+[fill=blue!66, draw=none] coordinates {(1,77.7) (2,85.8) (3,85)};
\addplot+[fill=red!66, draw=none] coordinates {(1,78.9) (2,86.4) (3,85.8)};
\legend{\small{IID}, \small{OOD}}; 
\end{axis} 
\end{tikzpicture} 
\hspace{0.2cm}
\begin{tikzpicture}
\begin{axis}[
title={(b)  Limited testing length},
xlabel={Testing buffer length},
ylabel={FG-ARI$(\%)$},
xtick={1,2,3,4}, 
xticklabels={Ball, B6, B4, B2}, 
width=0.34\textwidth,
height=4.2cm,
bar width=0.2cm,
y label style={at={(axis description cs:0.18,0.5)},anchor=south},
ymin=0, ymax=100,
axis lines=left,
enlarge x limits={abs=0.5},
legend style={draw=none,at={(1.35,0.8)},},
] 
\addplot+ coordinates {(1, 91.1) (2, 89.8) (3, 87.6) (4, 85.5)}; 
\addplot+ coordinates {(1, 69.1) (2, 67.3) (3, 66.6) (4, 61.3)}; 
\addplot+ coordinates {(1, 57.5) (2, 47.2) (3, 42.3) (4, 36.0)}; 
\addplot+ coordinates {(1, 40.9) (2, 15.9) (3, 13.3) (4, 10.5)}; 
\addplot+ coordinates {(1, 36.8) (2, 12.8) (3, 9.9) (4, 6.8)}; 
\legend{\small{MOVi-A}, \small{MOVi-B}, \small{MOVi-C}, \small{MOVi-D}, \small{MOVi-E}}; 
\end{axis} 
\end{tikzpicture} 
\hspace{0.2cm}
\begin{tikzpicture}
\begin{axis}[
ybar,
title={(c) Limited training length},
xlabel={Dataset},
ylabel={FG-ARI$(\%)$},
xtick={1,2}, 
xticklabels={MOVi-A, MOVi-E}, 
width=0.31\textwidth,
height=4.2cm,
y label style={at={(axis description cs:0.2,0.5)},anchor=south},
ymin=0, ymax=100,
axis lines=left,
enlarge x limits={abs=0.5},
legend pos=north west,
legend style={
    draw=none,%fill opacity=0,
    legend image code/.code={\fill[draw=black,fill=none] (0cm,-0.1cm) rectangle (0.2cm,0.2cm);},
    legend columns=4,  
    at={(0.2,1.12)},
},
] 
\addplot+ coordinates {(1, 92.3) (2, 17.9)}; 
\addplot+ coordinates {(1, 91.1) (2, 36.8)}; 
\addplot+ coordinates {(1, 90.4) (2, 30.1)}; 
\addplot+ coordinates {(1, 90.7) (2, 23.6)}; 
\legend{\small{B12}, \small{B6}, \small{B4}, \small{B2}}; 
\end{axis} 
\end{tikzpicture} 
\vspace{-6pt}
\caption{ (a) Results on out-of-distribution evaluation splits with new objects and backgrounds. (b) Ablation study of the buffer size on model performance during testing phase. The $x$-axis represents buffer sizes during testing. (c) Ablation study of the impact of buffer size on model performance during training phase. The different bars represent buffer sizes.}
% during training with  of 12, 6, 4, and 2 frames. }
\label{fig5}
\vspace{-3pt}
\end{figure*}
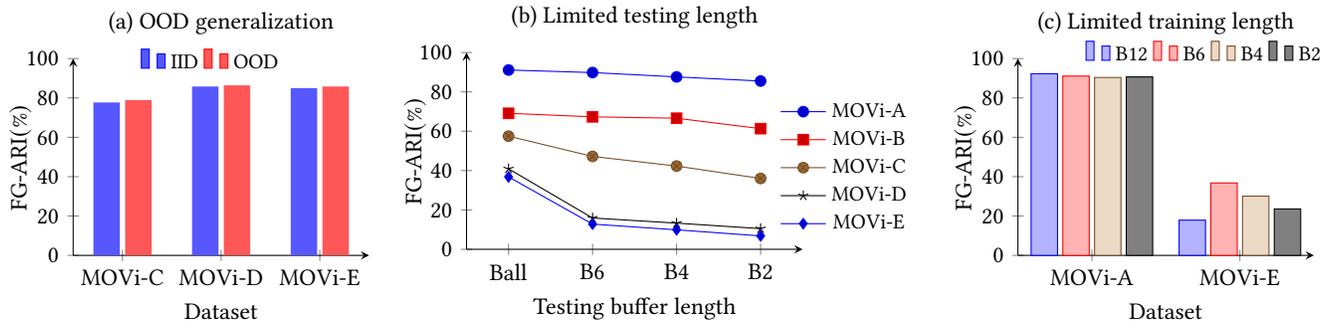

\begingroup
\setlength{\tabcolsep}{0.18em}
\begin{table*}[t!]
\renewcommand\arraystretch{1.1}
\caption{Ablation study of buffer on CLEVRER during the testing phase. `Perception' represents perceptual results of STATM-SAVi with different buffer sizes. `Prediction' shows prediction results of STATM trained with slots extracted by STATM-SAVi with different buffer sizes. `VQA' indicates our model's accuracy on predictive questions when using various buffer sizes.}
\vspace{-6pt}
\label{tab5}
\begin{center}
\begin{tabular}{lcclcclcc}
\toprule
\multicolumn{1}{l}{\bf \multirow{2}*{Model}} &\multicolumn{2}{c}{\bf Perception} &\multicolumn{1}{l}{} &\multicolumn{2}{c}{\bf Prediction} &\multicolumn{1}{l}{} &\multicolumn{2}{c}{\bf VQA} \\
\cline{2-3} \cline{5-6} \cline{8-9} 
{} &FG-ARI(\%)\quad &FG-mIoU(\%)\quad &{} &FG-ARI(\%)\quad &FG-mIoU(\%) &{}&per opt.(\%)\quad &per ques.(\%)\quad \\
\hline 
STATM (24)&93.5&77.6&{}&64.6&49.6&{}&96.62&93.63\\
STATM (32)&93.5&77.6&{}&64.5&49.5&{}&96.41&93.09\\
STATM (48)&93.5&77.6&{}&64.6&49.6&{}&96.07&92.67\\
STATM (128)&93.5&77.6&{}&64.2&49.3&{}&96.21&92.90\\
\bottomrule
\end{tabular}
%\vspace{-3pt}
\end{center}
\end{table*}
\endgroup

In this section, we aim to evaluate the influence of different components of STATM, using STATM-SAVi as a baseline. 

\textbf{Memory Buffer.}  We have designed two sets of experiments to evaluate the impact of the memory buffer: 1) In the first set, we allowed an unlimited memory buffer length during training, but restricted it to a fixed length during testing. 2) In the second set, we fixed the buffer length during training, and removed any buffer length restrictions during testing. To facilitate evaluation, we have not only assessed the model trained with 6 frames but also extended the training frames to 12. We show the results in Figure~\ref{fig5} (b) and (c). %For more results, please see Tables \ref{tab-a3} to \ref{tab-a5} in Appendix Section~\ref{append-ablation}.

\textit{Training phase}. From Figure \ref{fig5} (c), we observe that during the perception training phase, changes in buffer size have minimal impact on the model's performance on simple datasets. For complex datasets, initially increasing the buffer size improves model performance, but further increases eventually lead to a decline. Thus, during the perception training phase, buffer size  can be set to 6.

\textit{Testing phase}. Since the MOVi validation set contains only 24 frames, it does not adequately demonstrate the impact of memory size during the testing phase. Therefore, in Table \ref{tab5} perception column, we present evaluation results on CLEVRER dataset with a sequence of 128 frames during testing phase. We observe that increasing the buffer size during the testing phase initially improves perceptual outcomes, but beyond a certain size, it has negligible impact. Thus, we set the buffer size to 24 during the testing phase to avoid unnecessary computational costs.

Table \ref{tab5} prediction and VQA column display the effects of using different buffer sizes of STATM-SAVi to extract slots during the testing phase on prediction and VQA performance. It is observed that slots extracted by SAVi-STATM using different buffer sizes have almost no impact on STATM’s predictive capabilities. The impact of buffer size on VQA is minor, but beyond a certain threshold, VQA performance actually deteriorates. Therefore, for prediction and VQA tasks, we use 24 buffer size of STATM-SAVi to extract slot. %For more detailed results on the impact of buffer size on VQA, please refer to the Appendix Section~\ref{append-ablation} Table \ref{tab-d3}.

In summary, the buffer size can be set to 6 during the training of both the perception model and the long-term reasoning component. When testing the perception model, the buffer size should be set to 24. For the long-term reasoning component during use, a buffer size of 6 yields good results.

\begin{table}[t!]
\centering 
%\vskip 0.15in
\caption{Ablation study of perception for different spatiotemporal attention computation and fusion methods.}
\vspace{-10pt}
\label{tab6}
\begin{tabular}{lcclcc}
\toprule
\multicolumn{1}{l}{\bf \multirow{2}*{Model}} &\multicolumn{2}{c}{\bf mIoU$\uparrow(\%)$} &\multicolumn{1}{l}{} &\multicolumn{2}{c}{\bf FG-ARI$\uparrow(\%)$} \\
\cline{2-3} \cline{5-6} 
{} & A & E & {} & A & E\\
\hline 
STATM-SAVi (CS, ST) & 58.4&{-}&{}&90.9&{-} \\
STATM-SAVi (CS, TS) & 61.2&{-}&{}&89.7&{-} \\
STATM-SAVi (CS, T+S) & 67.5&8.5&{}&91.1&36.8 \\
STATM-SAVi (AS, T+S) & {-}&3.8&{}&{-}&12.2 \\
\bottomrule
\end{tabular}
\vspace{-6pt}
\end{table}

\textbf{Spatiotemporal Fusion and Computation.} In Table \ref{tab6}, we display results for STATM-SAVi using different spatiotemporal attention computation and fusion methods. From the first three rows of the table, it can be seen that the T+S spatiotemporal fusion method is relatively superior to both ST and TS. The last two rows indicate that the CS spatiotemporal attention structure is superior to AS. This may be due to the AS structure causing slot confusion from excessive slot interactions. Therefore, we use STAT with CS and T+S for our experiments by default.

\subsection{Limitations}
We did not assess our model using real-world datasets. Our perception and prediction models are trained separately, although they share a common predictive structure. In the future, we are more interested in implementing integrated training of our models.

\section{Conclusion}

In the real world, all objects follow the laws of physics. Intuitive physics serves as the bridge and connection through which humans comprehend the world. Our research aims to construct an object-centric system for object perception, learning of physics, and reasoning to explore whether deep learning models can learn physical concepts like humans, and use these learned physical laws to make inferences and predictions about the future motion of objects. We have designed a more reasonable prediction module called STATM, which clearly improved slot-based models in the context of scene understanding and prediction. We demonstrated that reasoning and prediction abilities influence each other. Through a series of experiments, we have shown the advantages of our model in tasks such as perception, prediction, and VQA. We also explore the impact of different spatiotemporal attention and fusion methods, and memory buffer on model perception and prediction. Although many challenges still remain in this field, the results presented in this paper illustrate that well-designed deep learning models can mimic human perception and prediction. In the future, we hope to implement joint training of our perception and prediction models, along with real-time perception and prediction, and validate the model in real-world scenarios.

\begin{acks}
The work is supported by the National Natural Science Foundation of China (No. 92270118, No. 62276269), the Beijing Natural Science Foundation (No. 1232009), and the Strategic Priority Research Program of the Chinese Academy of Sciences (No. XDB0620103). 
H. Sun and Y. Liu would like to thank the support from the Fundamental Research Funds for the Central Universities (No. 202230265, No. E2EG2202X2).
\end{acks}

\bibliographystyle{ACM-Reference-Format}
\balance
\bibliography{sample-base}


% \clearpage
\appendix

\setcounter{figure}{0}
\setcounter{table}{0}
\renewcommand{\thefigure}{S\arabic{figure}}
\renewcommand{\thetable}{S\arabic{table}}

\begin{center}
    {\Large \textbf{Appendix}}
\end{center}

This supplementary material file provides the appendix section to the main article.

%\section{Additional Parameter}
\begin{table*}[t!]
\renewcommand\arraystretch{1.1}
\caption{Comparison of the parameter number for different models.}
\label{tab-a1}
\vspace{-6pt}
\begin{center}
\begin{tabular}{lclc}
\toprule
\multicolumn{1}{c}{\bf Model} &\multicolumn{1}{c}{\bf Parameter Number} &\multicolumn{1}{c}{\bf Model} &\multicolumn{1}{c}{\bf Parameter Number} \\
\hline 
SAVi-Small & 895,268 & STATM-SAVi-Small & 961,572 \\
SAVi-Medium & 1,140,740 & STATM-SAVi-Medium & 1,207,044 \\
SAVi-Large & 22,273,412 & STATM-SAVi-Large & 22,339,716 \\
SAVi++ & 23,132,165 & STATM-SAVi++ & 23,264,389 \\
\bottomrule
\end{tabular}
\vspace{-3pt}
\end{center}
\end{table*}

\section{Baselines}
\label{append-baseline}

To validate the effectiveness of STATM, we chose three baselines which are state-of-the-art in object-centric video scene decomposition for improvement and comparison. All three models include a similar module: Slot Attention \cite{locatello2020object} followed by a predictor (a transformer encoder block \cite{vaswani2017attention}).

\textbf{SAVi.} The SAVi model \cite{kipf2021conditional} consists of five main components: encoder, decoder, slot initialization, corrector, and predictor. The encoder utilizes a convolutional neural network as a backbone to extract features from input video frames. The slot initialization is either a simple MLP (in the case of bounding boxes or center of mass coordinates) or a convolutional neural network (in the case of segmentation masks), responsible for initializing slots based on the conditioning information (bounding boxes, center of mass coordinates, or segmentation masks) in the first frame. The corrector employs Slot Attention \cite{locatello2020object} to update slot information based on visual features from the encoder. The predictor, a transformer encoder block \cite{vaswani2017attention}, utilizes self-attention among the slots for prediction. The output of the predictor initializes the corrector at the next time step, ensuring consistent object tracking over time. Finally, the decoder uses a Spatial Broadcast Decoder \cite{watters2019spatial} to generate RGB predictions of optical flow (or reconstructed frames) and an alpha mask.

For the SAVi model in perception, we chose the official implementation of SAVi with a small CNN as backbone, which is trained and evaluated on downscaled 64*64 frames. We initialized the slots in the first frame using bounding boxes as hints (slot initialization is a simple MLP). We selected optical flow as the target for training the model. During training, we adjusted the batch size to 32, while keeping other settings and parameters the same as SAVi-small.

\textbf{SAVi++.} SAVi++ \cite{elsayed2022SAVi++} has a structure similar to SAVi. During the training of SAVi++, we adjusted the batch size to 32, the number of training steps to 100k, while keeping other parameters consistent with described in SAVi++. For SAVi++$^{*}$, the hyperparameters are same as the official implementation.

\textbf{STEVE.} STEVE \cite{singh2022simple} is an unsupervised object-centric scene decomposition model. This baseline employs discrete VAE \cite{im2017denoising} for encoding and reconstructing input frames $x_t$ and generating discrete targets for the transformer decoder. It uses a similar structure combined with the encoder, corrector, and predictor in SAVi, called the recurrent slot encoder, to decompose input video frames $x_t$ into slots. The slot-transformer decoder uses the slots obtained from the recurrent slot encoder to learn to predict the sampling targets from discrete VAE by minimizing the cross-entropy loss. We make no modifications to the official implementation of STEVE.

\textbf {SlotFormer}. SlotFormer \cite{wu2023SlotFormer} is a transformer-based framework for object-centric visual simulation. It leverages slots extracted by upstream modules like SAVi to train a slot-based transformer encoder model for prediction purposes. Additionally, it utilizes results from rollout simulations to train Aloe \cite{ding2021attention} for Visual Question Answering (VQA) tasks. We employ an unsupervised approach to train SAVi-small on CLEVRER for slot extraction, with all other configurations consistent with the official settings. 

\textbf {SAVi-SlotFormer}. To evaluate whether enhancements to the predictor can alter the performance of SAVi, we replaced the predictor in SAVi with SlotFormer, which is currently the best-performing model in object-centric prediction, featuring a memory buffer and transformer encoders predictor. All other settings are consistent with SAVi.

\textbf {G-SWM}. G-SWM \cite{lin2020improving} is an unsupervised, object-centric predictive model that calculates foreground and background distributions through two separate modules, subsequently rendering and combining these results for dynamic prediction. Object interactions and occlusions are managed through a simple graph neural network. The official implementation has been trained using CLEVRER, yielding results comparable to those obtained with SlotFormer \cite{wu2023SlotFormer}.

\textbf {SAVi-dyn}. In SlotFormer \cite{wu2023SlotFormer}, Wu et al. enhanced prediction capabilities by replacing the Transformer predictor in SAVi with a Transformer-LSTM  module in PARTS \cite{zoran2021parts}. We trained SAVi-dyn using the same setup as theirs.

\textbf {DCL}. DCL \cite{chen2021grounding} utilizes a trajectory extractor to monitor each object over time, representing it as a latent, object-centric feature vector. Building on this foundational representation, DCL employs graph networks to learn and approximate the dynamic interactions among objects. 

% \textbf {VRDP}. VRDP \cite{ding2021dynamic} is designed to jointly learn visual concepts and infer physics models of objects and their interactions from both videos and language. It primarily consists of three modules: a visual perception module, a concept learner, and a differentiable physics engine. We have implemented VRDP with a visual perception module that is trained based on object properties.

\textbf {Aloe}. Aloe \cite{ding2021attention} trains transformers using slots for prediction. To enable a direct comparison with models such as SlotFormer, we use the Aloe model as re-implemented in the SlotFormer paper, ensuring all hyperparameters and settings are kept consistent with those described in the paper. 

\section{Additional Training Setup}
\label{append-trainsetup}

\textbf{Experiments of STATM-SAVi and STATM-SAVi++.} Referring to Section~\ref{exp-41}, we train our models (STATM-SAVi and STATM-SAVi++) for 100k steps with a batch size of 32 using Adam \cite{kingma2014adam}. Same as SAVi++ \cite{elsayed2022SAVi++}, we linearly increase the learning rate for 5000 steps to 0.0002 (starting from 0) and then decay the learning rate with a Cosine schedule \cite{loshchilov2016sgdr}. We split each video into sub-sequences of 6 frames to train the model (In the ablation study studying the impact of training stage buffer size on the model, sub-sequences are set to 4/6/12, referring to Section~\ref{exp-43}). In the initialization of slots for the first frame, bounding boxes are utilized as contextual cues. For MOVi-A, B, and C datasets, the number of slots is set to 11. In the case of datasets MOVi-D and E, the number of slots is set to 24. We use 1 iteration per frame for the Slot Attention \cite{locatello2020object} module. All other parameters and settings for each model remain consistent with their respective baselines. We implement models in JAX using the Flax neural network library, unless stated otherwise. 

\textbf{Experiments of STATM-SAVi++$^{*}$.} Further, we modify the training steps to 500k and change the batch size to 64 to train the STATM-SAVi++ model. Additionally, for the Slot Attention on the MOVi-E dataset, we adjust the number of iterations to 2 per frame. All other parameters and settings remain consistent with the SVAi++ \cite{elsayed2022SAVi++}. 

\textbf{Experiments of STATM-STEVE.}  Due to the necessity of memory in STATM, we modify the training subsequence length to 3/6/24 (corresponding to batch sizes of 24/12/8). During the experiments, we found that the STATM-STEVE model is sensitive to the MSE loss in discrete VAE \cite{im2017denoising}, and the addition of STATM slightly increases the difficulty of fitting of the model, which becomes more pronounced with an increase in the buffer size of STATM module. Therefore, for a better improvement and testing of the impact of STATM on STEVE, a two-stage training approach can be attempted, where a discrete VAE is first trained, followed by the training of STATM and other modules. All other parameters and settings remain consistent with the STEVE \cite{singh2022simple}. We implement STATM-STEVE in PyTorch. %The model is trained on 2*A100 with 80GB memory each, and 500 epochs require approximately 6 days. 

\textbf{Experiments of Prediction and VQA.} We train our models (STATM-SAVi) for 400k steps with a batch size of 64 on the CLEVRER dataset to extract slots. The number of slots is set to 7, with a learning rate of 0.0001. For prediction, we subsample the video by a factor of 2 to train STATM, conducting approximately 500k training steps with a batch size of 64 and a learning rate of 0.0002. We use rollout slots to train Aloe, targeting around 300k steps with a learning rate of 0.0001 and a batch size of 128. We use the Adam optimizer and apply the same warm-up and decay learning rate schedule for the first 2.5\% of the total training steps.

\section{Additional Parameter}
\label{append-parameter}

Using the STATM structure as a predictor does lead to a slight increase in the parameter count of the SAVi and SAVi++ models. However, under the same training settings, our model achieves superior metrics. This suggests that the moderate increase in the parameter count doesn't significantly increase the training complexity of our model.

STATM-SAVi-Small has a parameter increase of approximately 66K compared to SAVi-Small, which is notably smaller than the parameter increase seen in SAVi-Large compared to SAVi-Small (around 21378K parameters). Moreover, our STATM-SAVi-Small model, trained for 100k steps with a batch size of 32, performs similarly to the official SAVi-Large model, trained for 500k steps with a batch size of 64. This further highlights the reasonableness and superiority of our designed prediction module.

\section{Additional Experimental Results}
\label{append-result}

\textbf{Additional Metrics.} All models were trained in a conditional setting, initializing slots using ground-truth bounding box information in the first frame. Consequently, the results for STATM-SAVi++* in Table~\ref{tab1} are measured from the second frame onward, aligning with the evaluation method in SAVi++ \cite{elsayed2022SAVi++}. All other evaluation results include data from the first frame. 

%\textbf{Additional Segmentation Results.}
\begin{table*}[t!]
\renewcommand\arraystretch{1.1}
\caption{Segmentation results on the first 6 frames of the MOVi dataset.}
\label{tab-a2}
\vspace{-10pt}
\begin{center}
\begin{tabular}{lccccclccccc}
\toprule
\multicolumn{1}{l}{\bf \multirow{2}*{Model}} &\multicolumn{5}{c}{\bf mIoU$\uparrow(\%)$} &\multicolumn{1}{l}{} &\multicolumn{5}{c}{\bf FG-ARI$\uparrow(\%)$} \\
\cline{2-6} \cline{8-12} 
{} & A & B & C & D & E & {} & A & B & C & D & E \\
\hline 
SAVi-S & 66.9&49.3&29.7&13.9&8.3&{}&92.3&80.1&69.2&45.5&32.2 \\
STATM-SAVi-S & 71.0&51.6&43.5&21.9&12.5&{}&92.6&81.7&73.0&50.2&54.7 \\
\cline{1-12}
SAVi++ & 85.2&59.5&55.3&49.8&30.7&{}&97.2&86.3&83.9&87.1&88.2 \\
STATM-SAVi++ & 85.8&59.8&56.8&56.7&31.1&{}&97.2&86.6&83.9&89.2&88.6 \\
\bottomrule
\end{tabular}
\vspace{-6pt}
\end{center}
\end{table*}

\textbf{Additional Segmentation Results.} In order to better assess our model, we conducted an evaluation using the first 6 frames of the videos. Detailed results can be found in Table~\ref{tab-a2}. Referring to Table~\ref{tab1} in the main text, we can observe the following trends: on simple datasets, the decline in our model's object segmentation and tracking capabilities over extended time sequences is comparable to that of the baseline model. However, on complex datasets like MOVi-C, D, and E, the decrease in our model's performance is significantly less than that of the baseline model. This indicates that the STATM is more suitable for handling object segmentation and tracking tasks in longer-time sequences and complex environments. This finding further validates the effectiveness of our STATM model.

\begin{table*}[t!]
\renewcommand\arraystretch{1.1}
\caption{Accuracy on different questions and average results on CLEVRER.}
\vspace{-10pt}
\label{tab-d2}
\begin{center}
\begin{tabular}{lccclcclccc}
\toprule
\multicolumn{1}{l}{\bf \multirow{2}*{Model}} & \multicolumn{1}{c}{\bf \multirow{2}*{Descriptive}} & \multicolumn{2}{c}{\bf \multirow{1}*{Explanatory}} &\multicolumn{1}{l}{} & \multicolumn{2}{c}{\bf \multirow{1}*{Predictive}} &\multicolumn{1}{l}{} & \multicolumn{2}{c}{\bf \multirow{1}*{Counterfactual}} & \multicolumn{1}{l}{\bf \multirow{2}*{Average}} \\
\cline{3-4} \cline{6-7} \cline{9-10}  
{}&{}&per opt.&per ques.&{}&per opt.&per ques.&{}&per opt.&per ques.&{} \\
\hline 
DCL&90.70&89.58&82.82&{}&90.52&82.03&{}&80.38&46.52&75.52 \\
VRDP&93.40&96.30&91.94&{}&95.68&91.35&{}&94.83&84.29&90.24 \\
SlotFormer&95.17&98.04&94.79&{}&96.50&93.29&{}&90.63&73.78&89.26 \\
STATM(Ours)&95.22&98.15&95.04&{}&96.62&93.63&{}&90.57&73.90&89.44 \\
\bottomrule
\end{tabular}
\vspace{-6pt}
\end{center}
\end{table*}

\begin{figure*}[t!]
\begin{small}
    \centering    
    \begin{minipage}[t]{1.0\linewidth}
    \centering
    \rotatebox[origin=c]{90}{Video} \
        \begin{tabular}{@{\extracolsep{\fill}}c@{}@{\extracolsep{\fill}}}
        \includegraphics[width=0.95\linewidth]{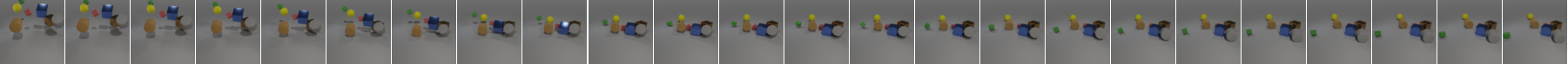}\\
        \end{tabular}
    \end{minipage}
    \begin{minipage}[t]{1.0\linewidth}
    \centering
    \rotatebox[origin=c]{90}{G.T.} \
        \begin{tabular}{@{\extracolsep{\fill}}c@{}@{\extracolsep{\fill}}}
        \includegraphics[width=0.95\linewidth]{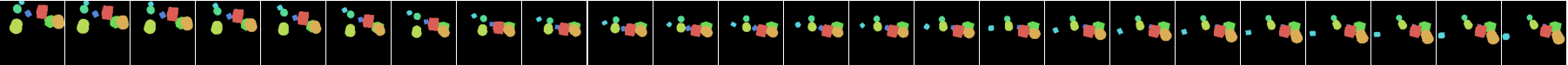}\\
        \end{tabular}
    \end{minipage}   
    \begin{minipage}[t]{1.0\linewidth}
    \rotatebox[origin=c]{90}{SAVi++} \
    \centering
        \begin{tabular}{@{\extracolsep{\fill}}c@{}@{\extracolsep{\fill}}}
        \includegraphics[width=0.95\linewidth]{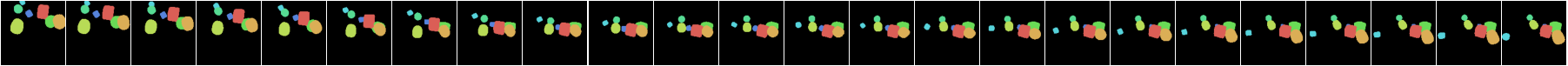}\\
        \end{tabular}
    \end{minipage}
    \begin{minipage}[t]{1.0\linewidth}
    \centering
    \rotatebox[origin=c]{90}{Ours++} \
        \begin{tabular}{@{\extracolsep{\fill}}c@{}@{\extracolsep{\fill}}}
        \includegraphics[width=0.95\linewidth]{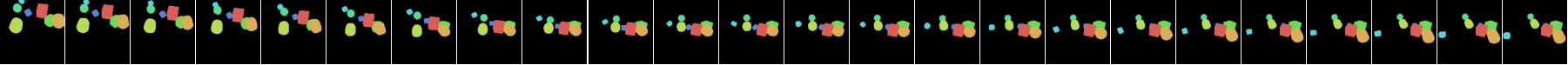}\\
        \end{tabular}
    \end{minipage}  
    \vspace{1pt} \\
    \begin{minipage}[t]{1.0\linewidth}
    \centering
    \rotatebox[origin=c]{90}{Video} \
        \begin{tabular}{@{\extracolsep{\fill}}c@{}@{\extracolsep{\fill}}}
        \includegraphics[width=0.95\linewidth]{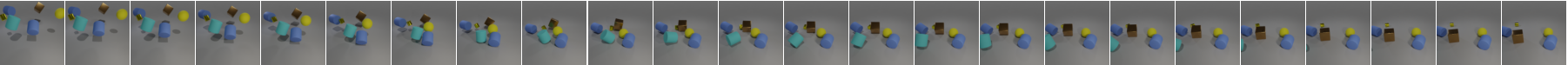}\\
        \end{tabular}
    \end{minipage}
    \begin{minipage}[t]{1.0\linewidth}
    \centering
    \rotatebox[origin=c]{90}{G.T.} \
        \begin{tabular}{@{\extracolsep{\fill}}c@{}@{\extracolsep{\fill}}}
        \includegraphics[width=0.95\linewidth]{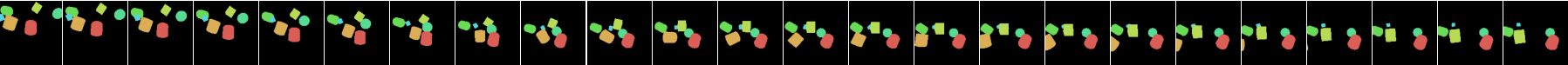}\\
        \end{tabular}
    \end{minipage}   
    \begin{minipage}[t]{1.0\linewidth}
    \rotatebox[origin=c]{90}{SAVi++} \
    \centering
        \begin{tabular}{@{\extracolsep{\fill}}c@{}@{\extracolsep{\fill}}}
        \includegraphics[width=0.95\linewidth]{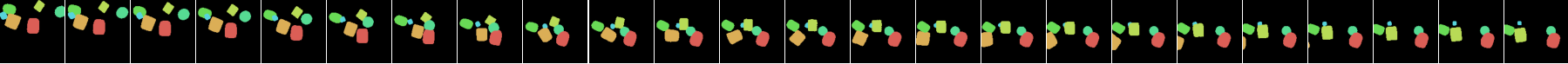}\\
        \end{tabular}
    \end{minipage}
    \begin{minipage}[t]{1.0\linewidth}
    \centering
    \rotatebox[origin=c]{90}{Ours++} \
        \begin{tabular}{@{\extracolsep{\fill}}c@{}@{\extracolsep{\fill}}}
        \includegraphics[width=0.95\linewidth]{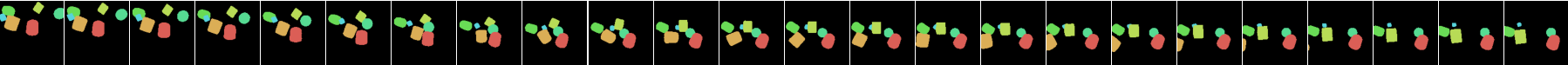}\\
        \end{tabular}
    \end{minipage}  
    \vspace{-10pt}
    \caption{Qualitative results of our model compared to SAVi++ on the MOVi-A dataset.}
\label{fig-a1}
\vspace{-6pt}
\end{small}
\end{figure*}

\begin{figure*}[t!]
\begin{small}
    \centering    
    \begin{minipage}[t]{1.0\linewidth}
    \centering
    \rotatebox[origin=c]{90}{Video} \
        \begin{tabular}{@{\extracolsep{\fill}}c@{}@{\extracolsep{\fill}}}
        \includegraphics[width=0.95\linewidth]{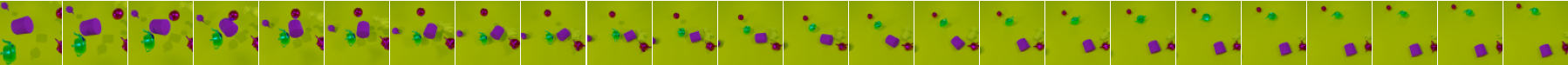}\\
        \end{tabular}
    \end{minipage}
    \begin{minipage}[t]{1.0\linewidth}
    \centering
    \rotatebox[origin=c]{90}{G.T.} \
        \begin{tabular}{@{\extracolsep{\fill}}c@{}@{\extracolsep{\fill}}}
        \includegraphics[width=0.95\linewidth]{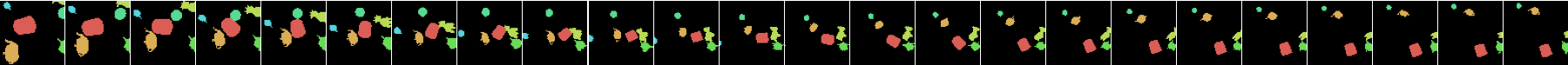}\\
        \end{tabular}
    \end{minipage}   
    \begin{minipage}[t]{1.0\linewidth}
    \rotatebox[origin=c]{90}{SAVi++} \
    \centering
        \begin{tabular}{@{\extracolsep{\fill}}c@{}@{\extracolsep{\fill}}}
        \includegraphics[width=0.95\linewidth]{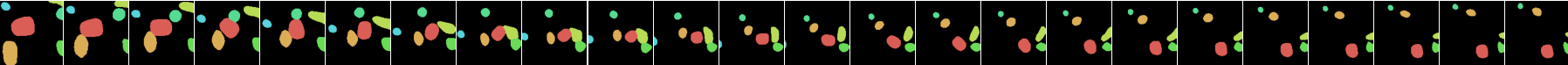}\\
        \end{tabular}
    \end{minipage}
    \begin{minipage}[t]{1.0\linewidth}
    \centering
    \rotatebox[origin=c]{90}{Ours++} \
        \begin{tabular}{@{\extracolsep{\fill}}c@{}@{\extracolsep{\fill}}}
        \includegraphics[width=0.95\linewidth]{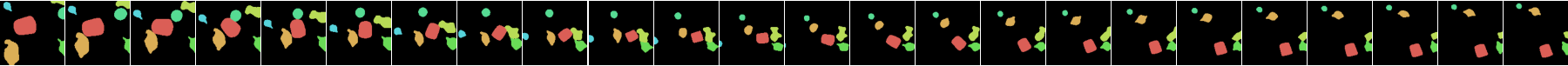}\\
        \end{tabular}
    \end{minipage}  
    \vspace{1pt} \\
    \begin{minipage}[t]{1.0\linewidth}
    \centering
    \rotatebox[origin=c]{90}{Video} \
        \begin{tabular}{@{\extracolsep{\fill}}c@{}@{\extracolsep{\fill}}}
        \includegraphics[width=0.95\linewidth]{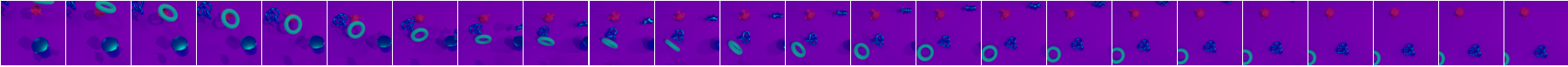}\\
        \end{tabular}
    \end{minipage}
    \begin{minipage}[t]{1.0\linewidth}
    \centering
    \rotatebox[origin=c]{90}{G.T.} \
        \begin{tabular}{@{\extracolsep{\fill}}c@{}@{\extracolsep{\fill}}}
        \includegraphics[width=0.95\linewidth]{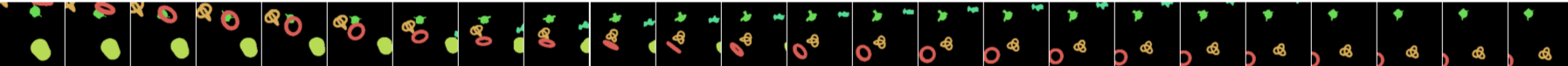}\\
        \end{tabular}
    \end{minipage}   
    \begin{minipage}[t]{1.0\linewidth}
    \rotatebox[origin=c]{90}{SAVi++} \
    \centering
        \begin{tabular}{@{\extracolsep{\fill}}c@{}@{\extracolsep{\fill}}}
        \includegraphics[width=0.95\linewidth]{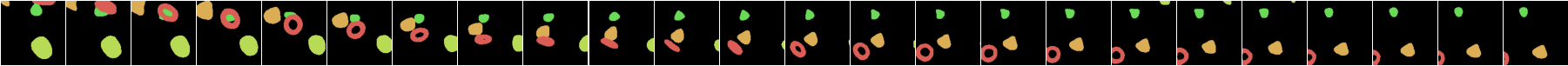}\\
        \end{tabular}
    \end{minipage}
    \begin{minipage}[t]{1.0\linewidth}
    \centering
    \rotatebox[origin=c]{90}{Ours++} \
        \begin{tabular}{@{\extracolsep{\fill}}c@{}@{\extracolsep{\fill}}}
        \includegraphics[width=0.95\linewidth]{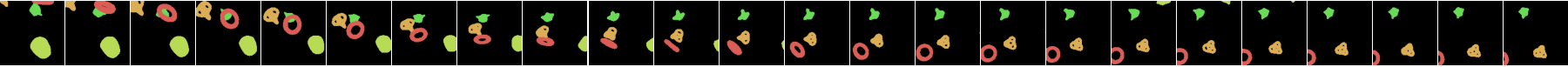}\\
        \end{tabular}
    \end{minipage}  
    \vspace{-10pt}
    \caption{Qualitative results of our model compared to SAVi++ on the MOVi-B dataset.}
\label{fig-a2}
%\vspace{-6pt}
\end{small}
\end{figure*}

\begin{figure*}[t!]
\begin{small}
    \centering    
    \begin{minipage}[t]{1.0\linewidth}
    \centering
    \rotatebox[origin=c]{90}{Video} \
        \begin{tabular}{@{\extracolsep{\fill}}c@{}@{\extracolsep{\fill}}}
        \includegraphics[width=0.95\linewidth]{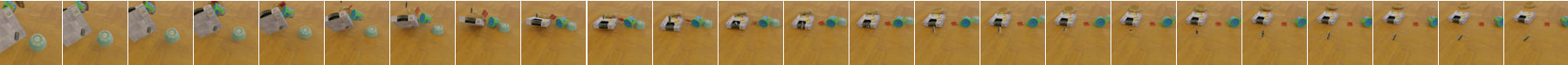}\\
        \end{tabular}
    \end{minipage}
    \begin{minipage}[t]{1.0\linewidth}
    \centering
    \rotatebox[origin=c]{90}{G.T.} \
        \begin{tabular}{@{\extracolsep{\fill}}c@{}@{\extracolsep{\fill}}}
        \includegraphics[width=0.95\linewidth]{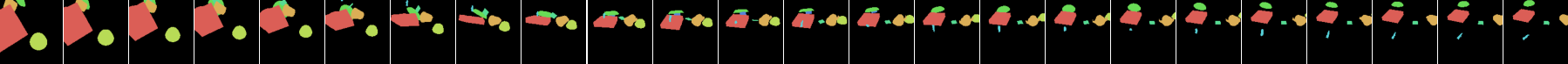}\\
        \end{tabular}
    \end{minipage}   
    \begin{minipage}[t]{1.0\linewidth}
    \rotatebox[origin=c]{90}{SAVi++} \
    \centering
        \begin{tabular}{@{\extracolsep{\fill}}c@{}@{\extracolsep{\fill}}}
        \includegraphics[width=0.95\linewidth]{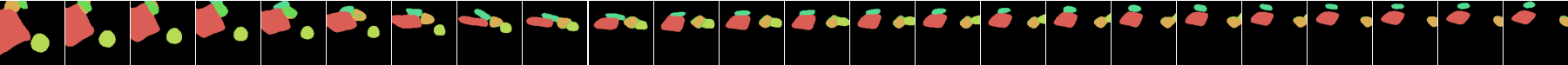}\\
        \end{tabular}
    \end{minipage}
    \begin{minipage}[t]{1.0\linewidth}
    \centering
    \rotatebox[origin=c]{90}{Ours++} \
        \begin{tabular}{@{\extracolsep{\fill}}c@{}@{\extracolsep{\fill}}}
        \includegraphics[width=0.95\linewidth]{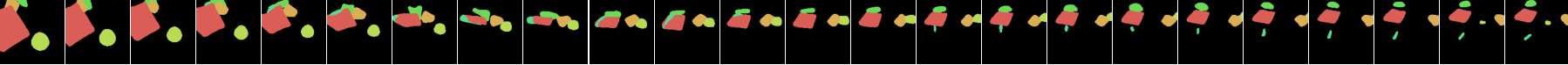}\\
        \end{tabular}
    \end{minipage}  
    \vspace{2pt} \\
    \begin{minipage}[t]{1.0\linewidth}
    \centering
    \rotatebox[origin=c]{90}{Video} \
        \begin{tabular}{@{\extracolsep{\fill}}c@{}@{\extracolsep{\fill}}}
        \includegraphics[width=0.95\linewidth]{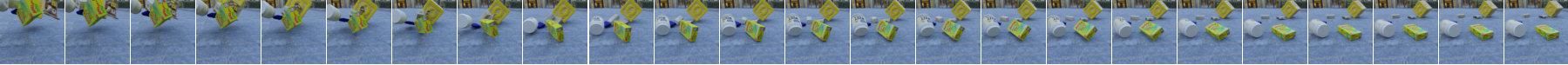}\\
        \end{tabular}
    \end{minipage}
    \begin{minipage}[t]{1.0\linewidth}
    \centering
    \rotatebox[origin=c]{90}{G.T.} \
        \begin{tabular}{@{\extracolsep{\fill}}c@{}@{\extracolsep{\fill}}}
        \includegraphics[width=0.95\linewidth]{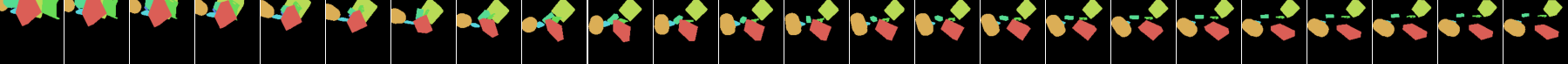}\\
        \end{tabular}
    \end{minipage}   
    \begin{minipage}[t]{1.0\linewidth}
    \rotatebox[origin=c]{90}{SAVi++} \
    \centering
        \begin{tabular}{@{\extracolsep{\fill}}c@{}@{\extracolsep{\fill}}}
        \includegraphics[width=0.95\linewidth]{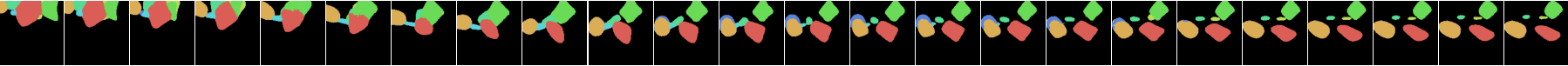}\\
        \end{tabular}
    \end{minipage}
    \begin{minipage}[t]{1.0\linewidth}
    \centering
    \rotatebox[origin=c]{90}{Ours++} \
        \begin{tabular}{@{\extracolsep{\fill}}c@{}@{\extracolsep{\fill}}}
        \includegraphics[width=0.95\linewidth]{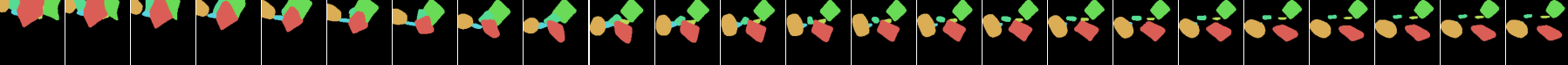}\\
        \end{tabular}
    \end{minipage} 
    \vspace{-10pt}
    \caption{Qualitative results of our model compared to SAVi++ on the MOVi-C dataset.}
\label{fig-a3}
\vspace{-6pt}
\end{small}
\end{figure*}

\begin{figure*}[t!]
\begin{small}
    \centering    
    \begin{minipage}[t]{1.0\linewidth}
    \centering
    \rotatebox[origin=c]{90}{Video} \
        \begin{tabular}{@{\extracolsep{\fill}}c@{}@{\extracolsep{\fill}}}
        \includegraphics[width=0.95\linewidth]{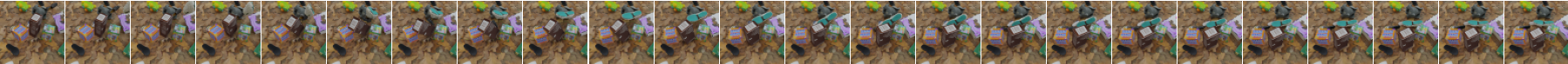}\\
        \end{tabular}
    \end{minipage}
    \begin{minipage}[t]{1.0\linewidth}
    \centering
    \rotatebox[origin=c]{90}{G.T.} \
        \begin{tabular}{@{\extracolsep{\fill}}c@{}@{\extracolsep{\fill}}}
        \includegraphics[width=0.95\linewidth]{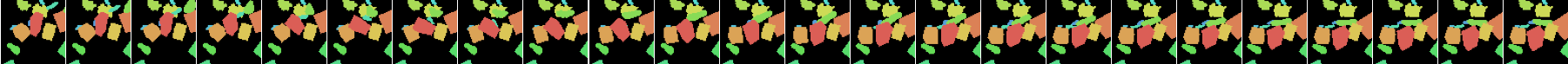}\\
        \end{tabular}
    \end{minipage}   
    \begin{minipage}[t]{1.0\linewidth}
    \rotatebox[origin=c]{90}{SAVi++} \
    \centering
        \begin{tabular}{@{\extracolsep{\fill}}c@{}@{\extracolsep{\fill}}}
        \includegraphics[width=0.95\linewidth]{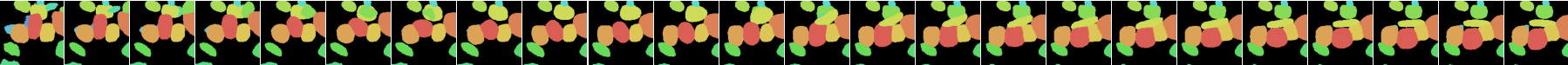}\\
        \end{tabular}
    \end{minipage}
    \begin{minipage}[t]{1.0\linewidth}
    \centering
    \rotatebox[origin=c]{90}{Ours++} \
        \begin{tabular}{@{\extracolsep{\fill}}c@{}@{\extracolsep{\fill}}}
        \includegraphics[width=0.95\linewidth]{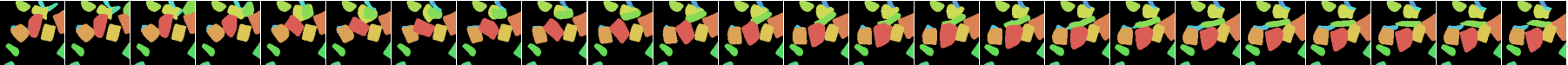}\\
        \end{tabular}
    \end{minipage}  
    \vspace{2pt} \\
    \begin{minipage}[t]{1.0\linewidth}
    \centering
    \rotatebox[origin=c]{90}{Video} \
        \begin{tabular}{@{\extracolsep{\fill}}c@{}@{\extracolsep{\fill}}}
        \includegraphics[width=0.95\linewidth]{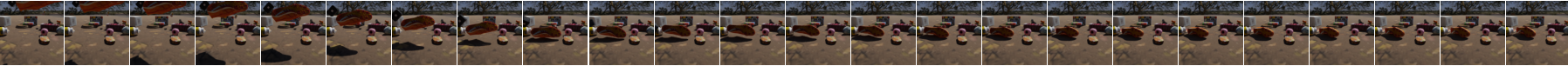}\\
        \end{tabular}
    \end{minipage}
    \begin{minipage}[t]{1.0\linewidth}
    \centering
    \rotatebox[origin=c]{90}{G.T.} \
        \begin{tabular}{@{\extracolsep{\fill}}c@{}@{\extracolsep{\fill}}}
        \includegraphics[width=0.95\linewidth]{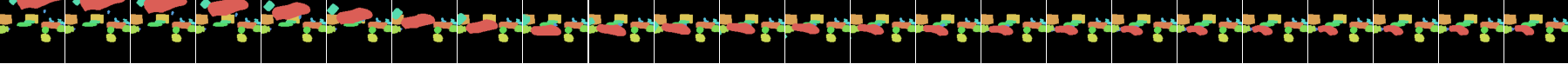}\\
        \end{tabular}
    \end{minipage}   
    \begin{minipage}[t]{1.0\linewidth}
    \rotatebox[origin=c]{90}{SAVi++} \
    \centering
        \begin{tabular}{@{\extracolsep{\fill}}c@{}@{\extracolsep{\fill}}}
        \includegraphics[width=0.95\linewidth]{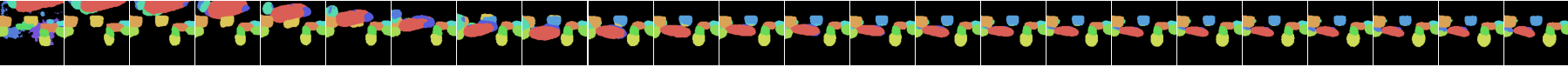}\\
        \end{tabular}
    \end{minipage}
    \begin{minipage}[t]{1.0\linewidth}
    \centering
    \rotatebox[origin=c]{90}{Ours++} \
        \begin{tabular}{@{\extracolsep{\fill}}c@{}@{\extracolsep{\fill}}}
        \includegraphics[width=0.95\linewidth]{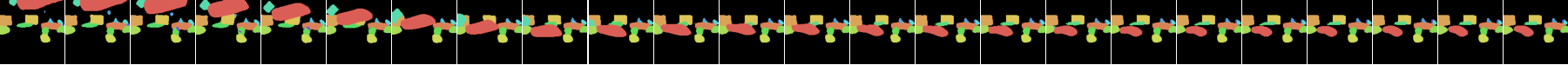}\\
        \end{tabular}
    \end{minipage}  
    \vspace{-10pt}
    \caption{Qualitative results of our model compared to SAVi++ on the MOVi-D dataset.}
\label{fig-a4}
%\vspace{-6pt}
\end{small}
\end{figure*}

\begin{figure*}[t!]
\begin{small}
    \centering    
    \begin{minipage}[t]{1.0\linewidth}
    \centering
    \rotatebox[origin=c]{90}{Video} \
        \begin{tabular}{@{\extracolsep{\fill}}c@{}@{\extracolsep{\fill}}}
        \includegraphics[width=0.95\linewidth]{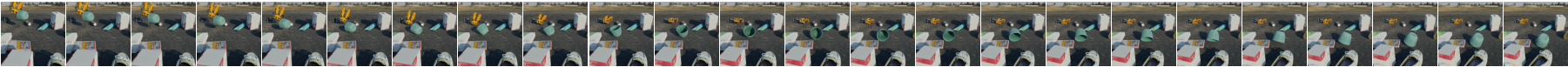}\\
        \end{tabular}
    \end{minipage}
    \begin{minipage}[t]{1.0\linewidth}
    \centering
    \rotatebox[origin=c]{90}{G.T.} \
        \begin{tabular}{@{\extracolsep{\fill}}c@{}@{\extracolsep{\fill}}}
        \includegraphics[width=0.95\linewidth]{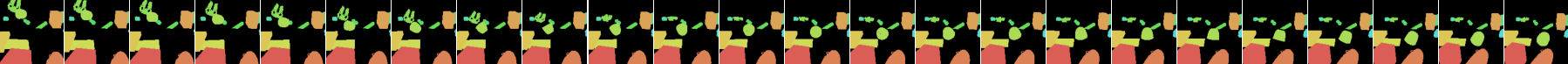}\\
        \end{tabular}
    \end{minipage}   
    \begin{minipage}[t]{1.0\linewidth}
    \rotatebox[origin=c]{90}{SAVi++} \
    \centering
        \begin{tabular}{@{\extracolsep{\fill}}c@{}@{\extracolsep{\fill}}}
        \includegraphics[width=0.95\linewidth]{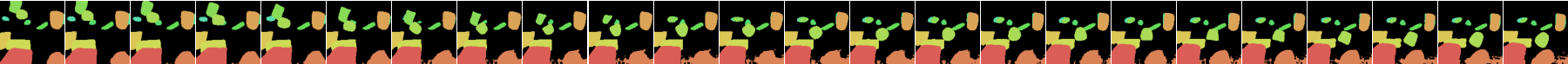}\\
        \end{tabular}
    \end{minipage}
    \begin{minipage}[t]{1.0\linewidth}
    \centering
    \rotatebox[origin=c]{90}{Ours++} \
        \begin{tabular}{@{\extracolsep{\fill}}c@{}@{\extracolsep{\fill}}}
        \includegraphics[width=0.95\linewidth]{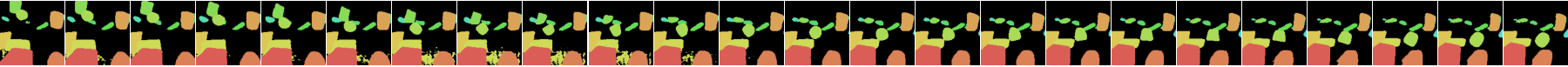}\\
        \end{tabular}
    \end{minipage}  
    \vspace{2pt} \\
    \begin{minipage}[t]{1.0\linewidth}
    \centering
    \rotatebox[origin=c]{90}{Video} \
        \begin{tabular}{@{\extracolsep{\fill}}c@{}@{\extracolsep{\fill}}}
        \includegraphics[width=0.95\linewidth]{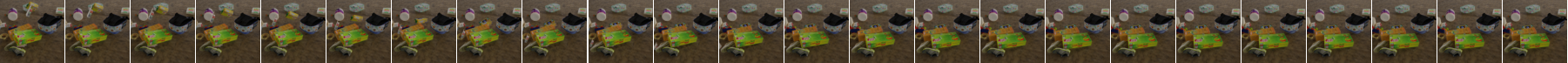}\\
        \end{tabular}
    \end{minipage}
    \begin{minipage}[t]{1.0\linewidth}
    \centering
    \rotatebox[origin=c]{90}{G.T.} \
        \begin{tabular}{@{\extracolsep{\fill}}c@{}@{\extracolsep{\fill}}}
        \includegraphics[width=0.95\linewidth]{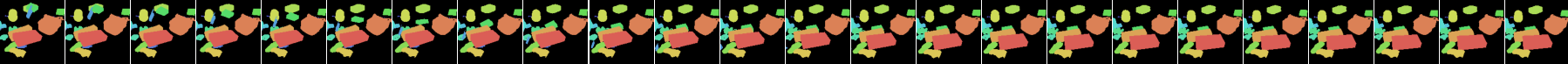}\\
        \end{tabular}
    \end{minipage}   
    \begin{minipage}[t]{1.0\linewidth}
    \rotatebox[origin=c]{90}{SAVi++} \
    \centering
        \begin{tabular}{@{\extracolsep{\fill}}c@{}@{\extracolsep{\fill}}}
        \includegraphics[width=0.95\linewidth]{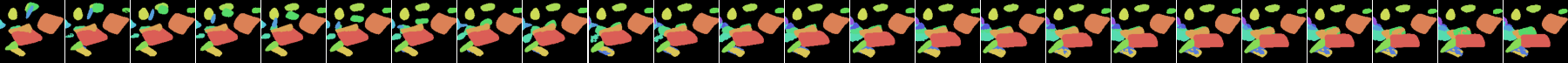}\\
        \end{tabular}
    \end{minipage}
    \begin{minipage}[t]{1.0\linewidth}
    \centering
    \rotatebox[origin=c]{90}{Ours++} \
        \begin{tabular}{@{\extracolsep{\fill}}c@{}@{\extracolsep{\fill}}}
        \includegraphics[width=0.95\linewidth]{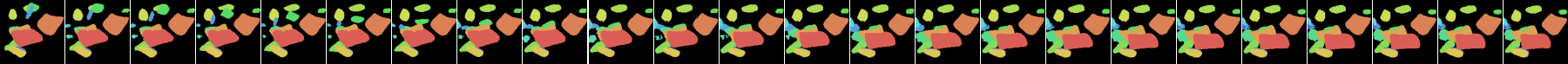}\\
        \end{tabular}
    \end{minipage}  
    \vspace{-10pt}
    \caption{Qualitative results of our model compared to SAVi++ on the MOVi-E dataset.}
\label{fig-a5}
%\vspace{-6pt}
\end{small}
\end{figure*}

\begin{figure*}[htbp]
\begin{center}
\centerline{\includegraphics[width=0.7\linewidth]{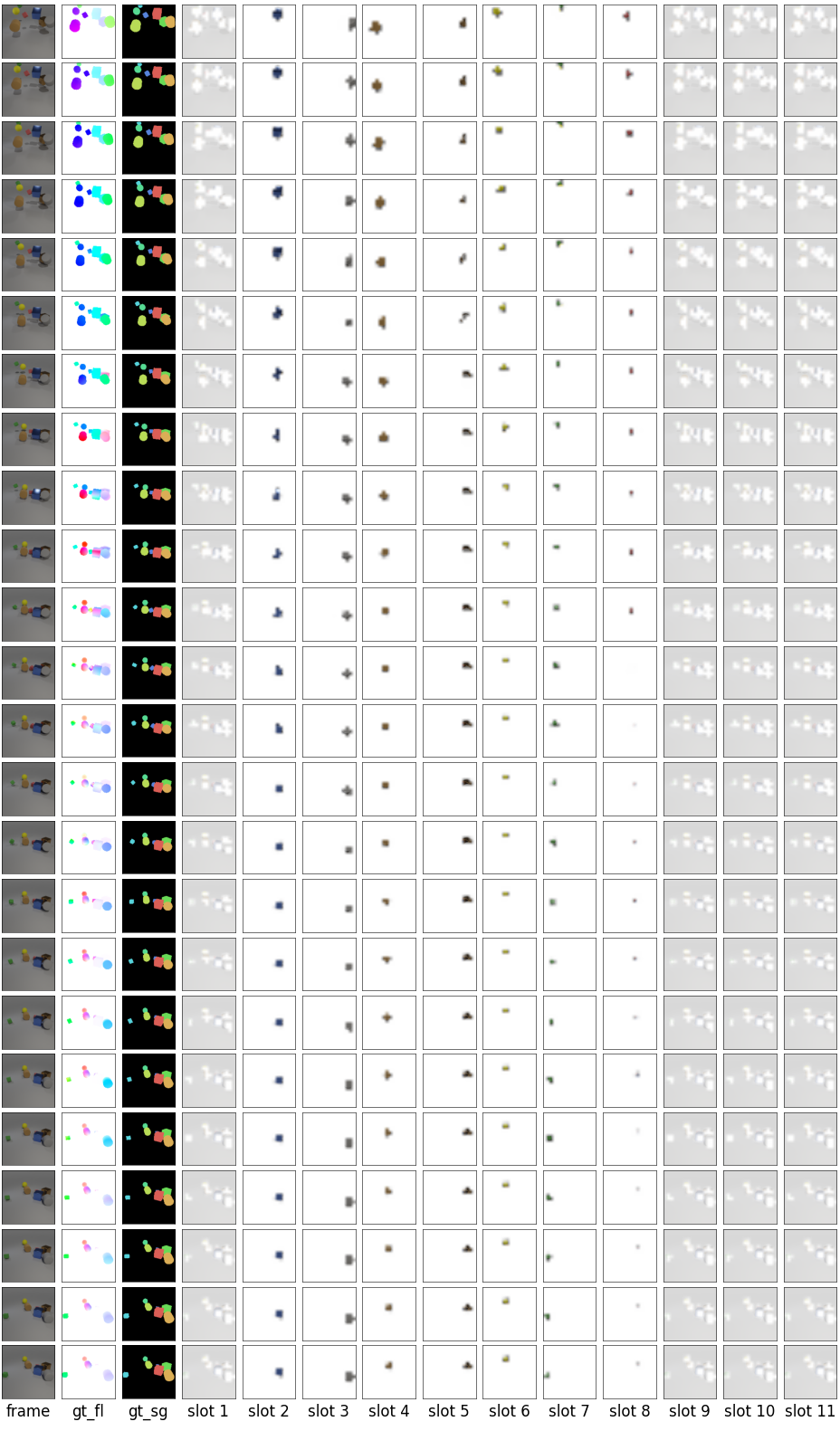}}
\vspace{-10pt}
\caption{Attention map visualization on the MOVi-A dataset.} 
\label{fig-b1}
\end{center}
\end{figure*}

\begin{figure*}[htbp]
\begin{center}
\centerline{\includegraphics[width=0.7\linewidth]{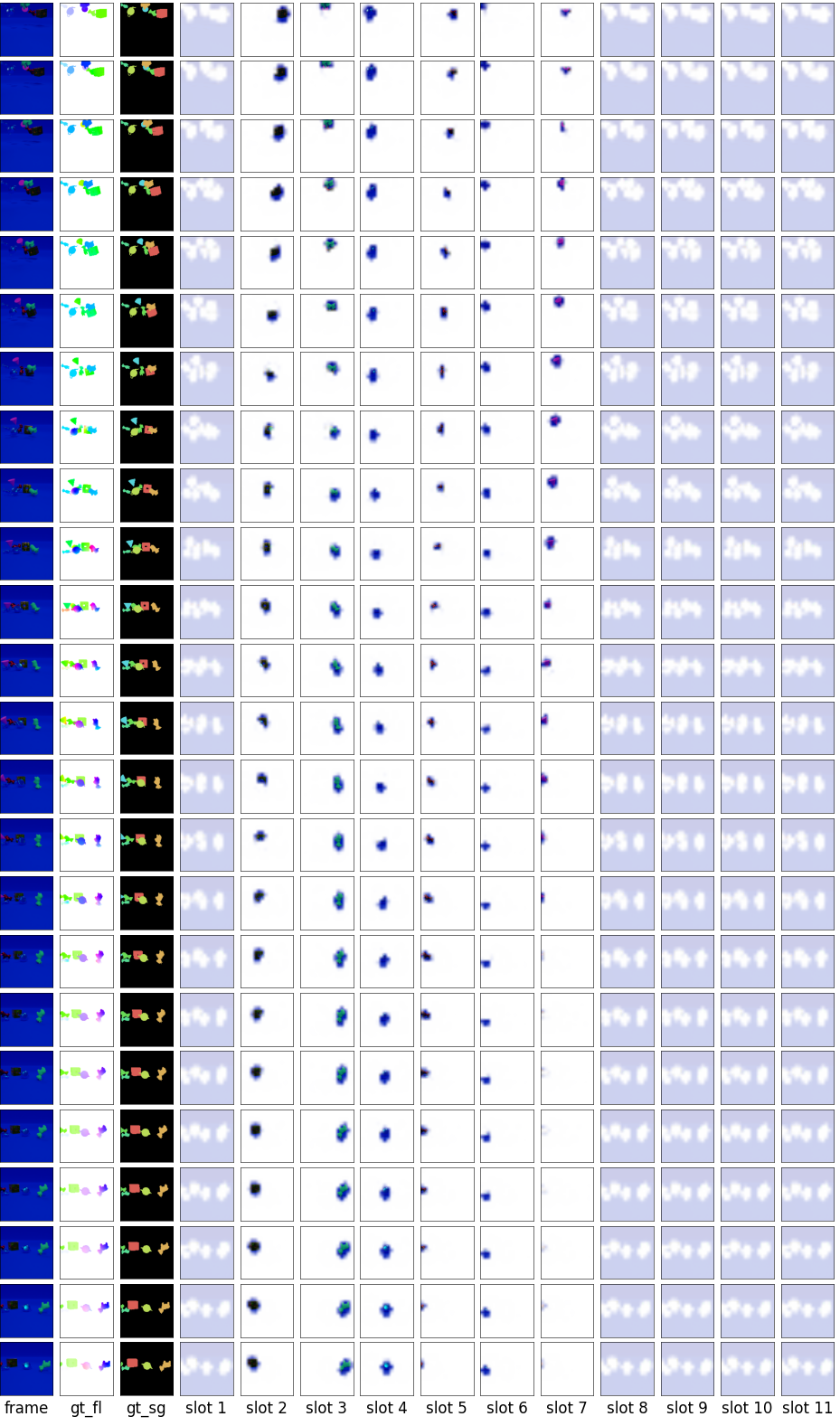}}
\vspace{-10pt}
\caption{Attention map visualization on the MOVi-B dataset.} 
\label{fig-b2}
\end{center}
\end{figure*}

\begin{figure*}[htbp]
\begin{center}
\centerline{\includegraphics[width=0.7\linewidth]{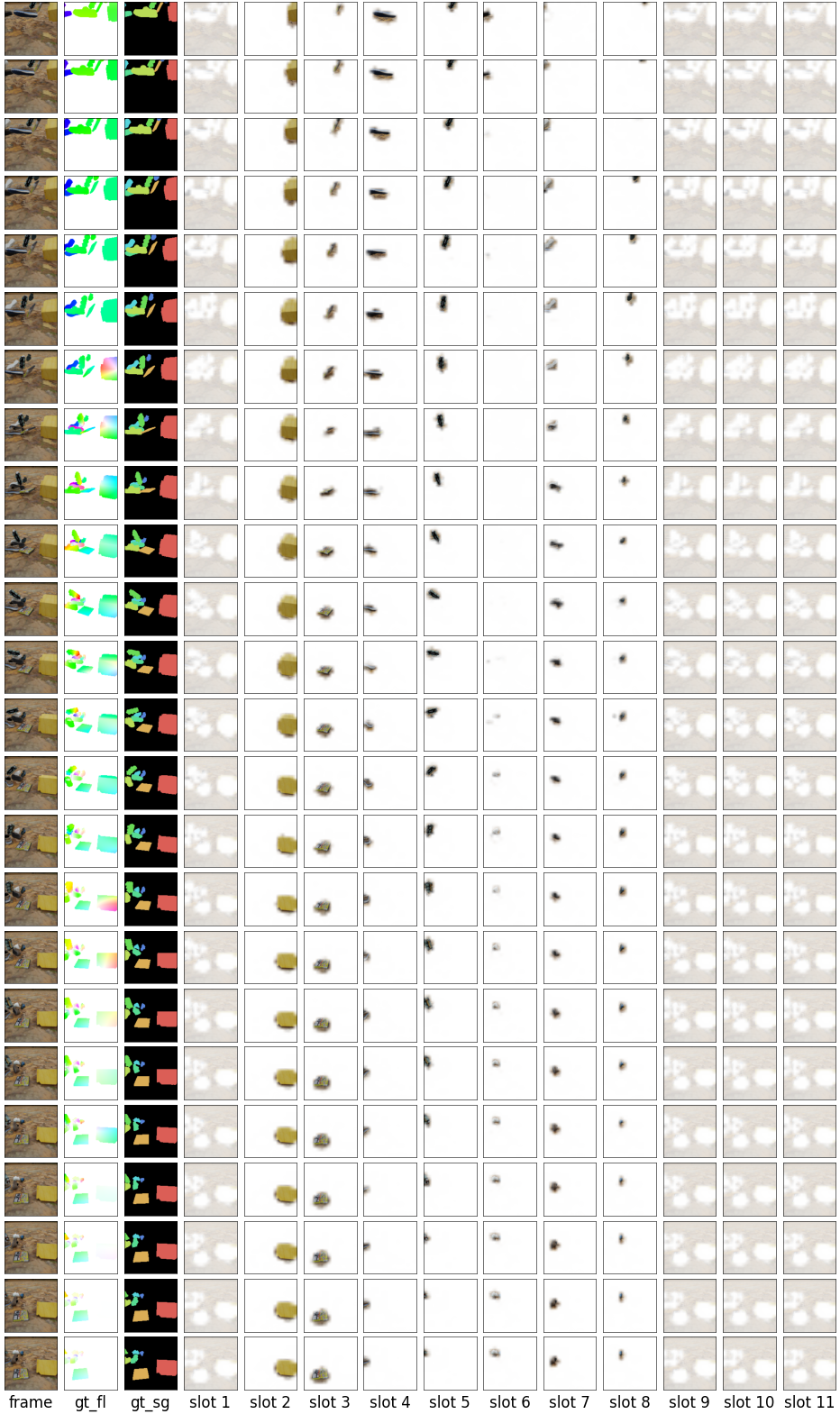}}
\vspace{-10pt}
\caption{Attention map visualization on the MOVi-C dataset.} 
\label{fig-b3}
\end{center}
\end{figure*}

\begin{figure*}[htbp]
\begin{center}
\centerline{\includegraphics[width=0.9\linewidth]{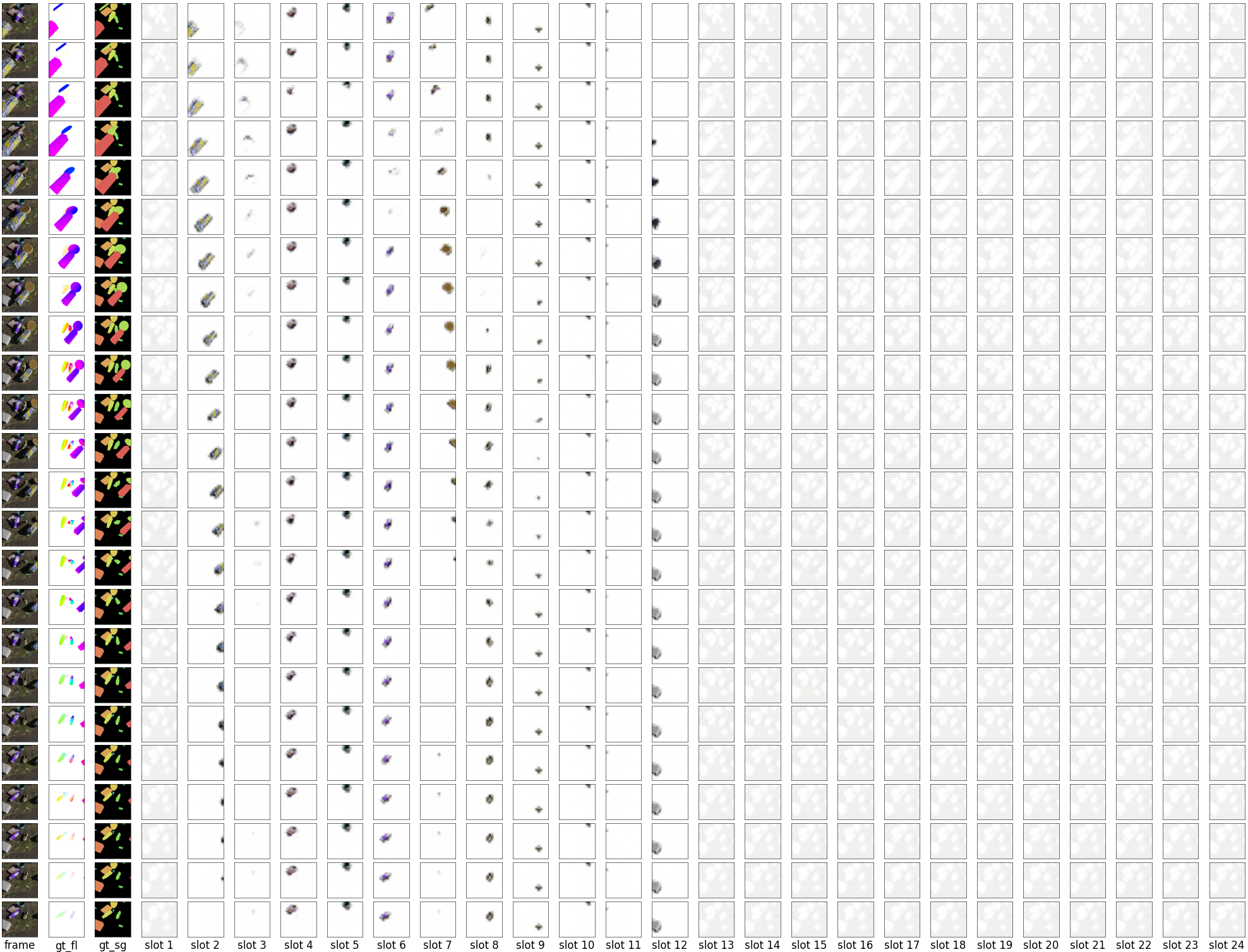}}
\vspace{-10pt}
\caption{Attention map visualization on the MOVi-D dataset.} 
\label{fig-b4}
\end{center}
\end{figure*}

\begin{figure*}[htbp]
\begin{center}
\centerline{\includegraphics[width=0.9\linewidth]{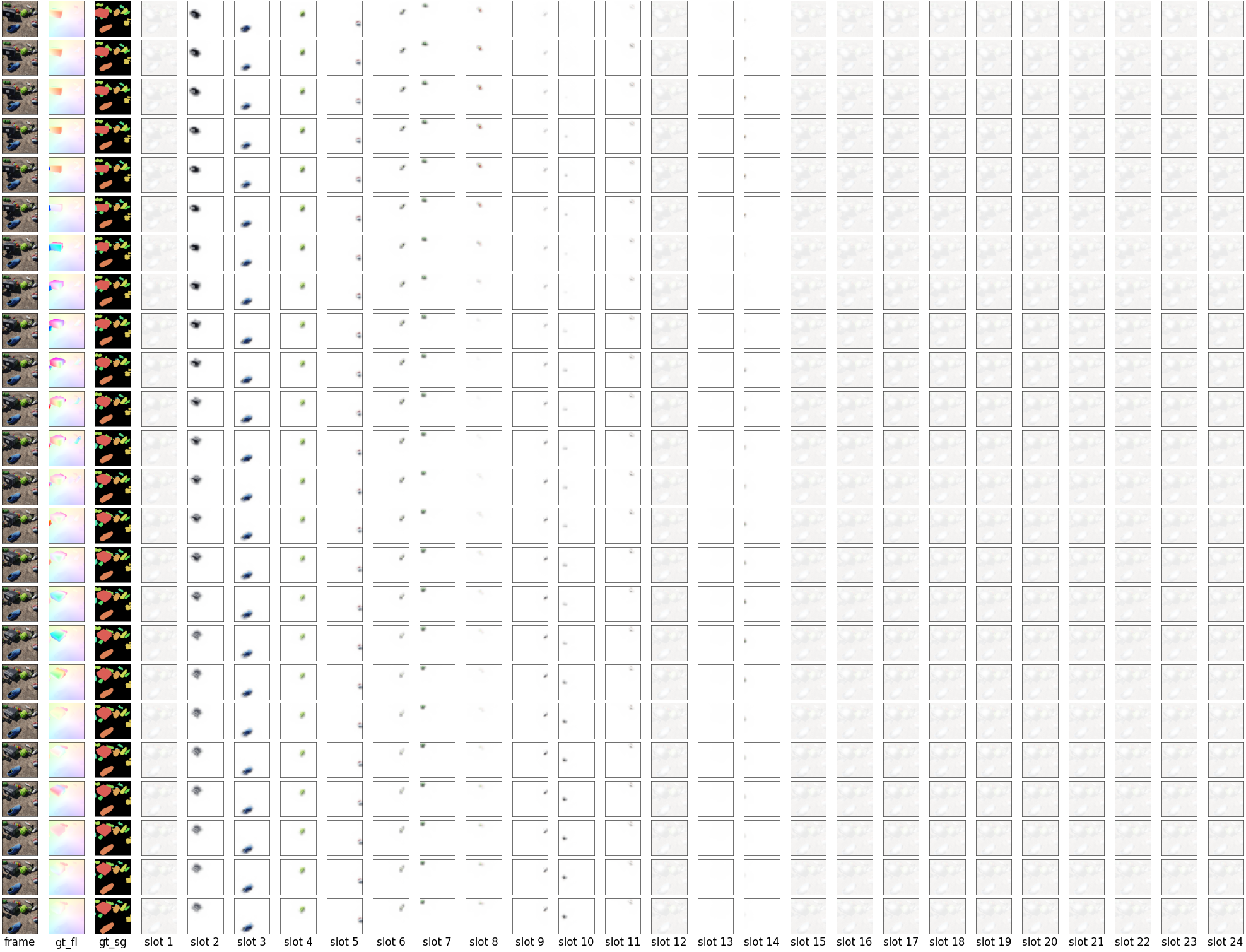}}
\vspace{-10pt}
\caption{Attention map visualization on the MOVi-E dataset.} 
\label{fig-b5}
\end{center}
\end{figure*}

\textbf{Additional Qualitative Results.} We show more qualitative results on longer time series in Figure~\ref{fig-a1} to Figure~\ref{fig-a5}. Meanwhile, To analyze slots and better illustrate the relationship between objects and slots, we visualized the attention map of the Slot Attention (corrector) in Figure~\ref{fig-b1} to Figure~\ref{fig-b5}.

\textbf{Additional VQA Results.} Detailed results about all questions on CLEVRER are presented in Table~\ref{tab-d2}.

\section{Additional Ablation Study}
\label{append-ablation}

\begin{table*}[t!]
\renewcommand\arraystretch{1.1}
\caption{Evaluation on all video frames of the model trained using 12 frames (B represents the size of the buffer during the testing phase).}
\vspace{-10pt}
\label{tab-a3}
\begin{center}
\begin{tabular}{lccccclccccc}
\toprule
\multicolumn{1}{l}{\bf \multirow{2}*{Model}} &\multicolumn{5}{c}{\bf mIoU$\uparrow(\%)$} &\multicolumn{1}{l}{} &\multicolumn{5}{c}{\bf FG-ARI$\uparrow(\%)$} \\
\cline{2-6} \cline{8-12} 
{} & A & B & C & D & E & {} & A & B & C & D & E \\
\hline 
%STATM (T6, Vall) & 67.5&42.7&34.0&17.0&8.5&{}&91.1&69.1&57.7&40.9&36.8 \\
%STATM (T6, V6) & 66.1&41.4&24.0&8.5&5.4&{}&89.8&67.3&47.2&15.9&12.8 \\
%STATM (T6, V4) & 64.2&40.6&20.6&7.4&4.9&{}&87.6&66.6&42.3&13.4&9.9 \\
%STATM (T6, V2) & 61.6&36.1&17.0&6.3&4.2&{}&85.5&61.3&36.0&10.5&6.8 \\
STATM (Ball) & 66.9&39.3&26.1&13.8&4.3&{}&92.3&72.9&62.5&59.6&17.9 \\
STATM (B12) & 66.2&39.3&25.9&13.2&3.9&{}&91.3&73.0&60.8&55.6&10.4 \\
STATM (B6) & 64.3&39.3&25.4&12.3&3.6&{}&89.3&72.7&57.4&50.6&5.6 \\
STATM (B4) & 62.8&39.1&24.8&11.8&3.4&{}&88.4&72.5&55.1&47.7&4.4 \\
STATM (B2) & 59.1&38.2&23.9&11.2&3.1&{}&85.5&70.6&51.1&44.0&3.5 \\
\bottomrule
\end{tabular}
\vspace{-6pt}
\end{center}
\end{table*}

\begin{table*}[t!]
\renewcommand\arraystretch{1.1}
\caption{Evaluation on the first 6 video frames of the models trained by 6 frames and 12 frames (T represents the number of frames used for training model, B represents the size of the buffer during the testing phase).}
\vspace{-10pt}
\label{tab-a4}
\begin{center}
\begin{tabular}{lccccclccccc}
\toprule
\multicolumn{1}{l}{\bf \multirow{2}*{Model}} &\multicolumn{5}{c}{\bf mIoU$\uparrow(\%)$} &\multicolumn{1}{l}{} &\multicolumn{5}{c}{\bf FG-ARI$\uparrow(\%)$} \\
\cline{2-6} \cline{8-12} 
{} & A & B & C & D & E & {} & A & B & C & D & E \\
\hline 
STATM (T6, Ball) & 71.0&51.6&43.5&21.9&12.5&{}&92.6&81.7&73.0&50.2&54.7 \\
STATM (T6, B6) & 71.0&51.6&43.5&21.9&12.5&{}&92.6&81.7&73.0&50.2&54.7 \\
STATM (T6, B4) & 71.0&51.3&42.7&19.7&12.0&{}&92.6&81.7&72.5&46.8&51.0 \\
STATM (T6, B2) & 70.8&49.2&38.7&14.9&10.2&{}&91.8&80.6&69.1&34.6&37.3 \\
STATM (T12, Ball) & 60.2&42.7&28.1&15.4&7.6&{}&92.7&82.9&73.6&55.5&33.5 \\
STATM (T12, B12) & 60.2&42.7&28.1&15.4&7.6&{}&92.7&82.9&73.6&55.5&33.5 \\
STATM (T12, B6) & 60.2&42.7&28.1&15.4&7.6&{}&92.7&82.9&73.6&55.5&33.5 \\
STATM (T12, B4) & 59.7&42.8&28.0&15.4&7.3&{}&92.1&82.9&73.3&54.5&29.3 \\
STATM (T12, B2) & 58.3&42.5&27.7&14.9&6.4&{}&89.6&82.5&71.6&50.2&19.4 \\
\bottomrule
\end{tabular}
\vspace{-6pt}
\end{center}
\end{table*}

\begin{table*}[t!]
\renewcommand\arraystretch{1.1}
\caption{Evaluation result of the model trained limited buffer length (T represents the size of the buffer during the training phase, B represents the size of the buffer during the testing phase).}
\vspace{-10pt}
\label{tab-a5}
\begin{center}
\begin{tabular}{lcccclcccc}
\toprule
\multicolumn{1}{l}{\bf \multirow{3}*{Model}} &\multicolumn{4}{c}{\bf mIoU$\uparrow(\%)$} &\multicolumn{1}{l}{} &\multicolumn{4}{c}{\bf FG-ARI$\uparrow(\%)$} \\
\cline{2-5} \cline{7-10} 
{} &\multicolumn{2}{c}{\bf First 6 frames} &\multicolumn{2}{c}{\bf All frames} &\multicolumn{1}{l}{} &\multicolumn{2}{c}{\bf First 6 frames} &\multicolumn{2}{c}{\bf All frames} \\
{} & A & E & A & E & {} & A & E & A & E \\
\hline 
STATM (T2, Ball) & 71.6&9.9&66.9&6.2&{}&92.6&41.2&90.7&23.6 \\
STATM (T2, B6) & 71.6&9.9&69.0&5.7&{}&92.6&41.2&91.5&22.7 \\
STATM (T2, B4) & 71.7&9.9&69.3&5.4&{}&92.6&41.5&91.2&20.0 \\
STATM (T2, B2) & 71.9&9.5&69.5&4.8&{}&92.6&41.3&91.2&15.5 \\
\hline 
STATM (T4, Ball) & 73.6&9.8&68.0&6.8&{}&92.5&41.7&90.4&30.1 \\
STATM (T4, B6) & 73.6&9.8&69.5&4.7&{}&92.5&41.7&90.3&15.1 \\
STATM (T4, B4) & 73.7&9.7&69.6&4.3&{}&92.4&41.3&90.1&11.8 \\
STATM (T4, B2) & 74.0&9.0&69.3&3.9&{}&92.1&38.4&89.2&9.1 \\
\hline 
STATM (T6, Ball) & 71.0&12.5&67.5&8.5&{}&92.6&54.7&91.1&36.8 \\
STATM (T6, B6) & 71.0&12.5&66.1&5.4&{}&92.6&54.7&89.8&12.8 \\
STATM (T6, B4) & 71.0&12.0&64.2&4.9&{}&92.6&51.0&87.6&9.9 \\
STATM (T6, B2) & 70.8&10.2&61.6&4.2&{}&91.8&37.3&85.5&6.8 \\
\bottomrule
\end{tabular}
\vspace{-6pt}
\end{center}
\end{table*}

\begin{table*}[t!]
\renewcommand\arraystretch{1.1}
\caption{Accuracy on different questions and average results of  different buffer sizes on CLEVRER.}
\vspace{-10pt}
\label{tab-d3}
\begin{center}
\begin{tabular}{lccclcclccc}
\toprule
\multicolumn{1}{l}{\bf \multirow{2}*{Model}} & \multicolumn{1}{c}{\bf \multirow{2}*{Descriptive}} & \multicolumn{2}{c}{\bf \multirow{1}*{Explanatory}} &\multicolumn{1}{l}{} & \multicolumn{2}{c}{\bf \multirow{1}*{Predictive}} &\multicolumn{1}{l}{} & \multicolumn{2}{c}{\bf \multirow{1}*{Counterfactual}} & \multicolumn{1}{l}{\bf \multirow{2}*{Average}} \\
\cline{3-4} \cline{6-7} \cline{9-10}  
{}&{}&per opt.&per ques.&{}&per opt.&per ques.&{}&per opt.&per ques.&{} \\
\hline 
STATM(24)&95.22&98.15&95.04&{}&96.62&93.63&{}&90.57&73.90&89.44\\
STATM(32)&95.34&98.23&95.30&{}&96.41&93.09&{}&90.91&74.69&89.61\\
STATM(48)&95.17&98.14&95.14&{}&96.07&92.67&{}&91.01&74.51&89.37\\
STATM(128)&95.15&97.84&94.15&{}&96.21&92.90&{}&90.50&73.44&88.91\\
\bottomrule
\end{tabular}
\vspace{-6pt}
\end{center}
\end{table*}

\textbf{Training with unlimited buffer length and testing with limited buffer length.} To better assess the impact of the buffer on the model, we trained the model using video sub-sequences of 12 frames, as shown in Table~\ref{tab-a3} and Table~\ref{tab-a4}. We observed that: 1) On relatively simple datasets like MOVi-A, B, C, and D dataset, increasing the amount of training data with additional information would enhance the model's segmentation capabilities. 2) Training with a buffer size of 12 results in a decrease in mIoU. SAVi++ augments the number of samples by utilizing sub-sequences of 6 frames. However, in this case, sub-sequences of 12 frames are employed, leading to a reduction in both the number of samples and hints. The decrease in both samples and hints may impact the model's ability to separate foreground and background, consequently causing a decline in mIoU. 3) On the MOVi-E dataset, increasing the number of training frames resulted in a decrease in the model's tracking and segmentation capabilities. This could be attributed to the limitations in the ability of the upstream modules to effectively extract image features. The findings from the SAVi and SAVi++, which used more powerful encoders and data augmentation to improve segmentation performance on MOVi-E, support this observation. Therefore, exploring the design of a more robust encoder and refining the corrector and guidance modules may yield unexpected improvements. We plan to further investigate this direction in future research.

\textbf{Training with limited buffer length and testing with limited buffer length.}  We intentionally limited the buffer size during both the training and testing phases, and the model evaluation results are presented in Table~\ref{tab-a5}. Remarkably, we found that the model trained with a smaller buffer experienced less impact from the buffer during the testing phase. For instance, consider a model trained with a buffer size of 2. When tested with a reduced buffer size on the MOVi-E dataset, the model experienced an approximately 8\% decrease in FG-ARI (from 23.6\% to 15.5\%). On the other hand, when testing with a reduced buffer size on the MOVi-E dataset, a model trained with a buffer size of 6 exhibited a FG-ARI decrease of about 30\% (from 36.8\% to 6.8\%). This has intriguing implications for the fusion of deep learning and cognitive science. However, it's important to note that real human learning and cognitive processes are likely more complex and influenced by various factors. This study provides a theoretical framework, but further theoretical substantiation and experimental validation are still needed.

\textbf{VQA results of different buffer size.} Detailed results for all questions on CLEVRER using different memory buffer sizes are presented in Table \ref{tab-d3}.

\section{STATM Structure in Different Model}
\label{append-structure}

Due to the differences between the SAVi/SAVi++ and STEVE models, there are certain distinctions in the enhanced models' STATM module as well. Table~\ref{tab-a6} illustrates the simplified algorithm for the STATM module in different models.

%\section{STATM Structure in Different Model}
\begin{table*}[t!]
\centering
\caption{Simplified algorithm for the STATM module in different models.}
\vspace{-10pt}
\label{tab-a6}
\begin{tabular}{cc}
\toprule
\multicolumn{1}{c}{\textbf{STATM in STATM-SAVi /SAVi++}} & \multicolumn{1}{c}{\textbf{STATM in STATM-STEVE}} \\
\midrule
\begin{minipage}{0.4\textwidth}
\vspace{0.1cm} 
\begin{algorithmic}
\STATE \textbf{Input:} $S_t$, $M_t$
\STATE
\STATE
\STATE $X_t$ = Spatiotemporal Attention($S_t$, $M_t$, $M_t$)
\STATE $X_t$ = LayerNorm($X_t$ + $S_t$)
\STATE $Y_t$ = MLP($X_t$)
\STATE $Y_t$ = LayerNorm($Y_t$)
\STATE \textbf{Return:} $X_t$ + $Y_t$ 
\end{algorithmic}
\vspace{0.1cm}
\end{minipage}
&
\begin{minipage}{0.4\textwidth}
\vspace{0.1cm} 
\begin{algorithmic}
\STATE \textbf{Input:} $S_t$, $M_t$
\STATE $S_t$ = LayerNorm($S_t$)
\STATE $M_t$ = LayerNorm($M_t$)
\STATE $X_t$ = Spatiotemporal Attention($S_t$, $M_t$, $M_t$)
\STATE $X_t$ = LayerNorm($X_t$ + $S_t$)
\STATE $Y_t$ = MLP($X_t$)
\STATE
\STATE \textbf{Return:} $S_t$ + $Y_t$ 
\end{algorithmic}
\vspace{0.1cm}
\end{minipage}
\\
\bottomrule
\end{tabular}
\vspace{-6pt}
\end{table*}

The STATM in STATM-SAVi /SAVi++ employs post-normalization, with the residual structure applies to the last MLP layer of the module. On the other hand, the STATM in STATM-STEVE utilizes pre-normalization, where $S_t$ and $M_t$ share normalization weights, and the entire module applies a residual structure. The size of the key($k$), query($q$), and value($v$) in the spatiotemporal attention for STATM-SAVi is 128, while in STATM-STEVE, it is 192.

%\textbf{Additional Qualitative Results.}

\end{document}